\documentclass{article}
\usepackage{graphicx}
\usepackage{subfigure}

\usepackage{natbib}

\usepackage{hyperref}

\usepackage[accepted]{icml2013}

\usepackage{graphicx}
\usepackage{amsmath}
\usepackage{amssymb,eucal}

\usepackage{mathrsfs}
\usepackage{amsfonts}
\usepackage{amsthm}
\usepackage{algorithmic,algorithm}

\def\T{{\!\top}}

\def\brho{ {\boldsymbol{\rho} } }

\def\bxi{ {\boldsymbol{\xi} } }
\def\bomega{ {\boldsymbol{\omega} } }
\def\bone{ {\bf 1 } }
\def\bzero{ {\bf 0 } }
\def\ba{ {\bf a } }
\def\bx{ {\bf x } }
\def\bp{ {\bf p } }
\def\bq{ {\bf q } }
\def\bu{ {\bf u } }

\def\bz{ {\bf z } }
\def\bv{ {\bf v } }
\def\bB{ {\bf b } }

\def\omega{w}
\def\preceq{\preccurlyeq}
\def\succeq{\succcurlyeq}

\icmltitlerunning{Learning Hash Functions Using Column Generation}

\usepackage{lipsum}
\newcommand\blfootnote[1]{%
  \begingroup
  \renewcommand\thefootnote{}\footnote{#1}%
  \addtocounter{footnote}{-1}%
  \endgroup
}

\renewcommand{\paragraph}{\textbf}

\expandafter\def\expandafter\normalsize\expandafter{%
\normalsize\setlength\abovedisplayskip{3pt}}

\expandafter\def\expandafter\normalsize\expandafter{%
\normalsize\setlength\belowdisplayskip{3pt}}

\begin{document}

\twocolumn[
\icmltitle{Learning Hash Functions Using Column Generation}

\icmlauthor{Xi Li$^*$}{xi.li03@adelaide.edu.au}\hspace{-.125cm}

\icmlauthor{Guosheng Lin$^*$}{guosheng.lin@adelaide.edu.au}

\icmlauthor{Chunhua Shen}{chunhua.shen@adelaide.edu.au}

\icmlauthor{Anton van den Hengel}{anton.vandenhengel@adelaide.edu.au}

\icmlauthor{Anthony Dick}{anthony.dick@adelaide.edu.au}
\icmladdress{Australian Centre for Visual Technologies, and School of
Computer Science, The University of Adelaide, Australia}

\icmlkeywords{Hashing, binary codes, column generation}

\vskip 0.3in
]

\begin{abstract}

    \blfootnote{* indicates equal contributions.}

    Fast nearest neighbor searching is becoming an increasingly
    important tool in solving many large-scale problems.  Recently a
    number of approaches to learning data-dependent hash functions
    have been developed.  In this
    work, we propose a column generation based method for learning
    data-dependent hash functions on the basis of proximity comparison
    information.
    Given a set of triplets that encode the pairwise proximity
    comparison information,
    our method learns hash functions that preserve the
    relative comparison relationships in the data as well as possible within the
    large-margin learning framework.  The learning procedure is
    implemented using column generation and hence is named CGHash. At
    each iteration of the column generation procedure, the best hash
    function is selected.  Unlike most other hashing methods, our
    method generalizes to new data points naturally; and has a training
    objective which is convex, thus ensuring that the global optimum
    can be identified.  Experiments demonstrate that the proposed method
    learns compact binary codes and that its retrieval performance
    compares favorably with state-of-the-art methods when tested on
    a  few benchmark datasets.

\end{abstract}

\section{Introduction}

    The explosive growth in the volume of data to be processed in
    applications such as web search and multimedia retrieval 
    increasingly demands fast similarity search and efficient data indexing/storage techniques. 
    Considerable
    effort has been spent on designing hashing methods which address
    both the issues of fast similarity search and
    efficient data storage 
    (for example,
    \cite{andoni2006near,weiss2008spectral,zhang2010self,norouzi2011minimal,kulis2009learning,gong2012iterative}).
    A hashing-based approach constructs a set of hash functions
    that map high-dimensional data samples to low-dimensional binary
    codes.  These binary codes can be easily loaded into the memory
    in order to allow rapid retrieval of data samples.  Moreover, the pairwise
    Hamming distance between these binary codes can be efficiently
    computed by using bit operations,
    which are well supported by modern processors, thus
    enabling efficient
    similarity calculation on large-scale datasets.
    Hash-based approaches have thus found a wide range of applications, including
    object recognition~\cite{torralba2008small}, information
    retrieval~\cite{zhang2010self}, local
    descriptor compression~\cite{ldahashing}, image
    matching \cite{korman2011coherency}, and many more.
    Recently a number of effective hashing methods have been
    developed which construct
    a variety of hash functions, mainly on the assumption that
    semantically similar data samples should have similar binary
    codes, such as random projection-based locality sensitive hashing
    (LSH)~\cite{andoni2006near}, boosting learning-based
    similarity sensitive coding (SSC) \cite{shakhnarovich2003fast},
    and  spectral hashing of \citet{weiss2008spectral} 
    which is inspired by Laplacian eigenmap. 

    In more detail, spectral hashing \cite{weiss2008spectral}
    optimizes a graph Laplacian based objective function such that
    in the learned low-dimensional binary space, 
    the local neighborhood structure of the original dataset is best
    preserved. 
    SSC \cite{shakhnarovich2003fast}
    makes use of  boosting to
    adaptively learn an embedding of the original space, 
    represented by a set of weak learners or hash functions.  This
    embedding aims to preserve the pairwise affinity relationships of
    training duplets (i.e., pairs of samples in the original space).
    These approaches have demonstrated that, in general,
    data-dependent hashing is superior to
    data-independent hashing with a typical example being LSH \cite{andoni2006near}.

    Following this vein, here we learn hash functions 
    using side information that is generally presented in a
    set of triplet-based constraints. 
    Note that the triples used for training can be generated 
    in an either supervised or unsupervised fashion.  
    The fundamental idea is to learn optimal hash functions such
    that, when using the learned weighted Hamming distance,  the relative
    distance comparisons of the form ``point $ \bx $ is closer to
    $ \bx^+$ than to $ \bx^- $'' are satisfied as well as possible ($\bx^+$ and $ \bx^-$ are respectively
    relevant and irrelevant samples to $ \bx$).
    This type of relative proximity comparisons have been successfully
    applied to learn quadratic distance metrics \cite{schultz2004learning,shen12jmlr}.
    Usually this type of proximity relationships do not require explicit
    class labels and thus are easier to obtain
    than either the class labels or the actual distances between data points.
    For instance, in content based image retrieval,
    to collect feedback, users may be required to report whether
    image $ \bx $ looks more similar to $ \bx^+$ than it is to a third image
    $ \bx^-$. This task  is typically much easier than to label each
    individual image.
    Formally,  we are given a set
    $\mathcal{C} = \{(\bx_i,
    \bx^{+}_i,\bx^{-}_i )|d(\bx_i,
    \bx^{+}_i)< d(\bx_i, \bx_i^{-})\}, i = 1,2,\cdots$, where
    $d(\cdot, \cdot)$ is some similarity measure (e.g., Euclidean
    distance in the original space; or semantic similarity measure
    provided by a user).
    As explained, {\em one may not explicitly know $d(\cdot, \cdot)$};
    instead, one may only be able to provide sparse proximity
    relationships.
    Using such a set of
    constraints, we formulate a learning problem in the large-margin
    framework.  By using a convex surrogate loss function,
    a convex optimization problem is obtained, but has an exponentially large number of
    variables. Column generation is thus employed to
    efficiently and optimally solve the formulated optimization
    problem.

    The main contribution of this work is to propose a novel hash function
    learning framework which has the following desirable properties.
        (i)
            The formulated optimization problem can be globally optimized.
    We show that  column generation can be used to iteratively find the
    optimal hash functions.
    The weights of all the selected hash functions for calculating
    the weighted Hamming distance are updated at each iteration.
        (ii) 
    The proposed framework is flexible and can accommodate various
    types of constraints.  We show how to learn hash functions
    based on proximity comparisons. Furthermore, the framework can
    accommodate different types of loss functions as well as
    regularization terms.
    Also, our hashing framework can use different types of
    hash functions such as linear functions,
    decision stumps/trees, RBF kernel functions, etc.

   \textbf{Related work}
    Loosely speaking, hashing methods may be categorized into two
    groups:
    data-independent and data-dependent.  Without using any training
    data, data-independent hashing methods usually generate a set of hash
    functions using randomization.  For instance,
    LSH of \citet{andoni2006near} use random projection and 
    thresholding to generate binary codes in the Hamming space, where
    the mutually close data samples in the Euclidean space are likely
    to have  similar binary codes.  Recently,  
    \citet{kulis2009kernelized} propose a kernelized version of
    LSH, which is capable of capturing the intrinsic relationships
    between data samples using kernels instead of linear inner
    products.
    In terms of learning methodology,
    data-dependent hashing methods
    can make use of unsupervised, supervised or semi-supervised
    learning techniques to learn a set of  hash
    functions that generate the compact binary codes.
    As for unsupervised learning,
    two typical approaches are used to obtain such compact binary
    codes, including thresholding the real-valued low-dimensional
    vectors (after dimensionality reduction)
    and direct optimization of a Hamming distance based objective
    function (e.g., spectral hashing \cite{weiss2008spectral},
    self-taught hashing \cite{zhang2010self}).
    The spectral hashing (SPH) method directly optimizes a graph
    Laplacian objective function in the Hamming space.
    Inspired by SPH, \citet{zhang2010self} developed the self-taught hashing
    (STH) method. At the first step of STH,  
     Laplacian graph embedding is used to generate
    a sequence of binary codes for each sample.  By viewing these
    binary codes as binary classification labels, a set of hash
    functions are obtained by training a set of bit-specific linear support vector
    machines.  \citet{liu2011hashingGraphs} proposed  a scalable
    graph-based hashing method which uses a   small-size   anchor graph to
    approximate the original neighborhood graph and alleviates the
    computational limitation of spectral hashing.

    As for the supervised learning case, a number of hashing methods
    take advantage of labeled training samples to build data-dependent
    hash functions.  These hashing methods often formulate
    hash function learning  as a classification problem. 
    For example, \citet{salakhutdinov2009semantic} proposed the restricted
    Boltzmann machine (RBM) hashing method using a multi-layer deep
    learning technique for binary code generation.
    \citet{ldahashing}
    use  Fisher linear discriminant analysis (LDA) to embed the
    original data samples into a lower-dimensional space, where the
    embedded data samples are binarized using thresholding.
    Boosting methods have also been employed to
    develop hashing methods such as
    SSC \cite{shakhnarovich2003fast} and Forgiving
    Hash~\cite{baluja2008learning}, both of which  learn
    a set of weak learners as hash functions in the boosting framework.
    It is demonstrated in~\cite{torralba2008small} that
    some data-dependent hashing methods like stacked RBM and
    boosting SSC perform much better than LSH on
    large-scale databases of millions of images.  
    \citet{wang2010semi} proposed a semi-supervised hashing method,
    which aims to ensure the smoothness of similar data
    samples and the separability of dissimilar data samples.  More
    recently, \citet{liu2012supervised} introduced a
    kernel-based supervised hashing method, where the hashing
    functions are nonlinear kernel functions. 

    The closest work to ours might be boosting based SSC hashing
    \cite{shakhnarovich2003fast}, which also learns a set of weighted
    hash functions through boosting learning.
    Ours differs SSC in the learning procedure. The resulting
    optimization problem of our CGHash is based on the concept of
    margin maximization. We have derived a meaningful Lagrange dual
    problem such that column generation can be applied to solve the
    semi-infinite optimization problem. In contrast, SSC is built on
    the learning procedure of AdaBoost, which employs stage-wise
    coordinate-descent optimization.  
    The weights associated with selected hash functions
    (corresponding weak classifiers in AdaBoost) are not fully updated at each iteration.
    Also the information used for training is different. We have used
    distance comparison information and SSC uses pairwise information.
    In addition, our work can accommodate various types of  constraints,
    and can flexibly adapt to different types of loss functions as
    well as regularization terms.
    It is unclear, for example,  how SSC can accommodate different  types 
    regularization that may encode useful prior information. 
    In this sense our CGHash is much more flexible.
    Next, we present our main results.

\section{The proposed algorithm}

    Given a set of  training samples $\bx_{m} \in  \mathcal{R}^{D}$,
    ($ m =1,2, \dots $), we aim to learn a set of hash functions
    $h_{j}(\bx) \in \cal H$, $ j = 1,2,\dots \ell,$ for mapping these
    training samples to a low-dimensional binary space, being
    described by a set of binary codewords
    $\mathbf{b}_{ m },( m =1,2,\dots).$ Here each $ {\bf b}_m $ is an
    $\ell$-dimensional binary vectors.  In the low-dimensional binary
    space, the codewords $\mathbf{b}_{m }$'s are supposed to preserve
    the underlying proximity information of corresponding $ \bx_i $'s
    in the original high-dimensional space.  Next we learn such
    hash functions $\{h_{j}(\bx)\}_{j=1}^{\ell}$ within the
    large-margin learning framework.

    Formally, suppose that we are given a set of triplets
    $\{(\bx_{i}, \bx^{+}_{i},\bx^{-}_{i})\}_{i=1}^{|\mathcal{I}|}$
    with
    $ \bx_{i}, \bx^{+}_{i}, \bx^{-}_{i} \in \mathcal{R}^{D}$
    and $\mathcal{I}$ being the triplet index set.
    These triplets encode the proximity comparison information
    such that the distance/dissimilarity
    between $\bx_{i}$ and ${\bx}^+_{i} $ is smaller than that
    between $\bx_{i}$ and ${\bx}^-_{i} $.
    Now we need to define the weighted Hamming distance
    for the learned binary codes:
$
        d_{\mathcal{H}}(\bx, \bz) =
        \sum_{j=1}^{\ell}\omega_{j}|h_{j}(\bx)-h_{j}(\bz)|,
$
where $\omega_{j}$ is a non-negative weight factor associated with the
$j$-th hash function.
    In our experiments, we have generated the triplets set as: $
    \bx_{i} $ and $ \bx^+_{i}$ belong to the same class and $ \bx_{i}
    $ and $ \bx^-_{i}$ belong to different classes. As discussed,
    these triplets may be sparsely provided by users in applications
    such as image retrieval.
    So we want the constraints
    $ d_{\mathcal{H}}( \bx_{i}, \bx^+_{i} ) < d_{\mathcal{H}}( \bx_{i}, \bx^-_{i} ) $ to be satisfied as
well as possible.
For notational simplicity, we define
$a^{[i]}_{j}=|h_{j}(\bx_{i})-h_{j}(\bx^-_{i})|-|h_{j}(\bx_{i})-h_{j}(\bx^+_{i})|$
and $d_{\mathcal{H}}( \bx_{i}, \bx^-_{i}) - d_{\mathcal{H}}( \bx_{i}, \bx^+_{i} )$
$ = $ $\bomega^{\T}\ba_{i}$ with  %
\begin{equation}
\ba_{i} = [a^{[i]}_{1}, a^{[i]}_{2},
\ldots, a^{[i]}_{\ell} ]^{\T}.
\label{EQ:A}  %
\end{equation}
In what follows, we describe the details of
our hashing algorithm using different types of
convex loss functions and regularization norms.
In theory,  any convex loss and regularization can be used in our
hashing  framework.
More details of our hashing  algorithm can be found
in Algorithm~\ref{alg:boosting_hash} and the supplementary
file \cite{shen_web}.

\subsection{Learning hashing functions with the hinge loss}

{\bf Hashing with $l_1$ norm regularization}
Using the hinge loss, we define the following large-margin
optimization problem: %
\begin{equation}
\begin{array}{cl}
\underset{\bomega, \bxi}{\min} &  \sum_{i=1}^{|\mathcal{I}|}\xi_{i} + C\|\bomega\|_{1}\\
\mbox{s.t.} &  \bomega \succeq \bzero,
\bxi \succeq \bzero; 
\\
& d_{\mathcal{H}}( \bx_{i}, \bx^-_{i}) - d_{\mathcal{H}}( \bx_{i}, \bx^+_{i} ) \geq 1 - \xi_{i},  \thickspace \forall i,\\
\end{array}
\label{eq:Hinge_Loss_Optimization_original} %
\end{equation}
where $\|\cdot\|_{1}$ is the 1-norm, $\bomega = (\omega_{1},
\omega_{2}, \ldots, \omega_{\ell})^{\T}$ is the weight vector;
$\bxi$ is the slack variable; $C$ is a  parameter
controlling the trade-off between the training error and model capacity,
and the symbol `$ \succeq $' indicates element-wise inequalities.
The optimization problem~\eqref{eq:Hinge_Loss_Optimization_original}
can be rewritten as: 
\begin{equation}
\begin{array}{cl}
\underset{\bomega, \bxi}{\min} & \sum_{i=1}^{|\mathcal{I}|}\xi_{i} + C\bone^{\T}\bomega\\
\mbox{s.t.} &  \bomega \succeq \bzero; \thickspace  \ba_{i}^{\T} \bomega \geq 1 - \xi_{i}, \thickspace \xi_{i}\geq 0, \thickspace \forall i,\\
\end{array}
\label{eq:Hinge_Loss_Optimization_variant} 
\end{equation}
where $\bone$ is the all-one column vector.
The corresponding dual problem is: %
\begin{equation}
\begin{array}{c}
\underset{\bu}{\max} \thickspace \bone^{\T}\bu, \;\;
\mbox{s.t.}  \thickspace A\bu \preceq C\bone, \thinspace \bzero \preceq \bu \preceq \bone,
\end{array}
\label{eq:dual_problem} 
\end{equation}
where the matrix $A = ( \ba_{1}, \ba_{2}, \ldots, \ba_{|\mathcal{I}|})  \in
\mathcal{R}^{ \ell
\times |\cal I| } $ and 
the symbol `$ \preceq $' indicates element-wise inequalities.

{\bf Hashing with $l_\infty$ norm regularization}
The primal problem
is formulated as: 
\begin{equation*}
\begin{array}{cl}
\underset{\bomega, \bxi}{\min} &  \sum_{i=1}^{|\mathcal{I}|}\xi_{i} 
+ C \| \bomega \|_{ \infty }
\\
\mbox{s.t.} &  \bomega \succeq \bzero, 
\ba_{i} ^\T \bomega   \geq 1 - \xi_{i},
\xi_i \geq 0, \forall i.\\
\end{array}
\end{equation*}
We can make the $ l_\infty $ regularization a constraint, 
\begin{equation}
\begin{array}{cl}
\underset{\bomega, \bxi}{\min} &  \sum_{i=1}^{|\mathcal{I}|}\xi_{i} \\
\mbox{s.t.} &  \bomega \succeq \bzero, \|\bomega\|_{\infty} \leq C';
  \ba_{i} ^\T \bomega
\geq 1 - \xi_{i},   \xi_{i} \geq 0, 
\thickspace \forall i.\\
\end{array}
\label{eq:Hinge_Loss_Optimization_original_L_infinity} 
\end{equation}

The dual form of the above optimization problem is: 
\begin{equation}
\begin{array}{cl}
\underset{\bu, \bq}{\min} & - \bone^{\T} \bu + C'\bone^{\T}\bq
\\
\mbox{s.t.} & A\bu \preceq \bq, \thinspace \bzero \preceq \bu \preceq \bone,
\end{array}
\label{eq:dual_problem_infinity_2} %
\end{equation}
where $C'$ is a positive constant.

\subsection{Hashing with a general convex loss function}

    Here we derive the  algorithm for learning hash functions with
    general convex loss. We assume that the general convex loss function
    $ f ( \cdot ) $ is
    smooth (exponential, logistic, squared hinge loss etc.)
    although our algorithm can be easily extended to non-smooth loss
    functions.

\vline
\begin{algorithm}[t!]
    \begin{minipage}[ctb]{8cm}
\caption{\footnotesize Hashing using column generation}
\label{alg:boosting_hash}
{                        \footnotesize
{\bf Input:}
{
    Training triplets $\{ (\bx_{i}, \bx_{i}^+, \bx_{i}^-) \}, i =
    1,2\cdots$
    and $ \ell $, the number of hash functions.
}

{\bf Output:}
{
 Learned hash functions $\{h_{j}(\bx)\}_{j=1}^{\ell}$ and the
 associated  weights $\bomega $.
}

    {\bf Initialize:} $\bu \leftarrow \frac{1}{ |\mathcal{I} | }$.

$ \;\;$ {\bf  for} $j = 1 $ to $ \ell$ {\bf do}
\vspace{-2mm}
\begin{enumerate}
        \setlength{\itemsep}{-1mm}
\item  Find the best hash function $h_{j}(\cdot)$ by solving the
sub-problem \eqref{eq:column_generation}.
\item Add $h_{j}(\cdot)$ to the hash function set;
\item 
    Update  $ \ba_i$, $
\forall i$ as in \eqref{EQ:A};
\item  Solve the primal problem for $ \bomega $ (using LBFGS-B
\cite{lbfgs})
and obtain
the dual variable $\bu$ using KKT condition \eqref{EQ:KKT1}.
\end{enumerate}
\vspace{-3mm}
$ \;\;$ {\bf     endfor}
}
\end{minipage}
\end{algorithm}

{\bf Hashing with $l_1$ norm regularization}
    Assume that we want to find a set of hash functions such that
    the set of constraints
    $
        d_{\mathcal{H}}( \bx_{i}, \bx^-_{i}) - d_{\mathcal{H}}( \bx_{i}, \bx^+_{i} )
    $
    $ = \bomega^{\T}\ba_{i} > 0, i = 1,2\dots $ hold as well as
    possible. These constraints do not have to be all strictly satisfied.
    Now, we need to define
    the margin $ \rho_i =  \bomega^{\T}\ba_{i}  $, and we want to
    maximize the margin with regularization.
    Using $l_1$ norm regularization to control the capacity, we may define
    the primal optimization problem as: %
\begin{equation}
\underset{\bomega, \brho}{\min}
\sum_{i=1}^{|\mathcal{I}|}f(\rho_{i}) + C\|\bomega\|_{1},
\mbox{s.t.} \;  \bomega \succeq \bzero; \rho_{i} =
\ba_{i}^{\T}\bomega, \thickspace \forall i.
\label{eq:General_Loss_Optimization_original} %
\end{equation}
    Here $f(\cdot)$ is a smooth convex loss function;
    $\bomega = [ \omega_{1}, \omega_{2}, \ldots, \omega_{\ell} ]^{\T}$
    is the weight vector that we are interested in optimizing.
     $C$ is a  parameter
    controlling the trade-off between the training error and model capacity.

    Also without this regularization, one can always make  $ \bomega $
    arbitrarily large
    to make the convex loss approach zero when all constraints are satisfied.
    Here because the possibility of hash functions can be extremely
    large or even infinite, we are not able to directly solve the
    problem \eqref{eq:General_Loss_Optimization_original}.
    We can use the column generation technique to iteratively
    and approximately solve the  original problem.
    Column generation is a technique originally used for large scale
    linear programming problems. 
    \citet{demiriz2002linear} used this method
    to design boosting algorithms.  At each iteration, one column---a
    variable in the primal or a constraint in the dual problem---is added when
    solving the restricted problem. Till one can not find any column
    violating the constraint in the dual, the solution of the
    restricted problem is identical to the optimal solution.
    Here we only need to obtain an approximate solution and in order
    to learn compact codes, we only care about the first few (e.g,
    60) selected hash functions.
    In theory, if we run the column generation  with a
    sufficient number of iterations, one can obtain a sufficiently
    accurate solution (up to a preset precision or no more hash
    functions can be found to improve the solution).

    We need to derive a meaningful Lagrange dual in order to use
    column generation.
    The Lagrangian is:
\begin{align*}
L  = \sum_{i=1}^{|\mathcal{I}|}f(\rho_{i}) +
C\bone^{\T}\bomega - \bp^{\T}\bomega +
\sum_{i=1}^{|\mathcal{I}|}u_{i}(\ba_{i}^{\T}\bomega - \rho_{i})\\
 = C\bone^{\T}\bomega - \bp^{\T}\bomega + \sum_{i=1}^{|\mathcal{I}|}u_{i}(\ba_{i}^{\T}\bomega)
- (\bu^{\T}\brho -  \sum_{i=1}^{|\mathcal{I}|}f(\rho_{i})),
\end{align*}
    where $\bp \succeq {\boldsymbol 0} $ and $\bu $ are Lagrange
    multipliers.
With the definition of Fenchel conjugate \cite{boyd}, we have the following
    relation:

$
\underset{ \brho}{\inf} \thinspace {L} 
$ 
$
    = - \underset{\brho}{\sup} 
    $
    $
    \thinspace (\bu^{\T}\brho -
$
$
    {\textstyle \sum}_{i=1}^{|\mathcal{I}|}f(\rho_{i}))
$
$
    = - {\textstyle \sum}_{i=1}^{|\mathcal{I}|}f^{\ast}(u_i)
$
and in order to have a finite infimum,  $ C\bone - \bp + A\bu =
\boldsymbol 0 $ must hold.
So we have $\bp \succeq 0$,  $A\bu \succeq -C\bone$.
 Here  the matrix
 $ A $ is defined in \eqref{eq:dual_problem}.  

Consequently, the corresponding dual problem
of~\eqref{eq:General_Loss_Optimization_original} can be written as:
\begin{equation}
\begin{array}{cl}
\underset{\bu}{\min} & \sum_{i=1}^{|\mathcal{I}|}f^{\ast}(u_i),
\mbox{s.t.}  \;
A\bu \succeq -C\bone.
\end{array}
\label{eq:General_Loss_Optimization_dual1A} \vspace{-0.1cm}
\end{equation}
Here $f^*(\cdot) $ is the Fenchel conjugate of $ f(\cdot) $.
By reversing the sign of $\bu$, we can reformulate \eqref{eq:General_Loss_Optimization_dual1A}
as its equivalent form:
\begin{equation}
    \underset{\bu}{\min} \quad
    {\textstyle \sum}_{i=1}^{|\mathcal{I}|}f^{\ast}(-u_i),
    \,
    \mbox{s.t.} \;
    A\bu \preceq C\bone.
\label{eq:General_Loss_Optimization_dual1B} \vspace{-0.1cm}
\end{equation}
    Since we assume that $ f(\cdot) $ is smooth, the
    Karush-Kuhn-Tucker (KKT) condition establishes the connection
    between \eqref{eq:General_Loss_Optimization_dual1B} and
    \eqref{eq:General_Loss_Optimization_original} at optimality:
    \begin{equation}
        \label{EQ:KKT1}
        u_i^\star = - f'( \rho_i^\star ), \forall i.
    \end{equation}
    In other words, the dual variable is determined by the gradient of
    the loss function in the primal. 
    So if we solve the primal problem
    \eqref{eq:General_Loss_Optimization_original},
    from the primal solution $ \bomega^\star $, we can calculate
    the dual solution $ \bu^\star $ using \eqref{EQ:KKT1}. But
    the other way around may not be true.

\begin{figure*}[t]
\centering
\includegraphics[scale=0.216]{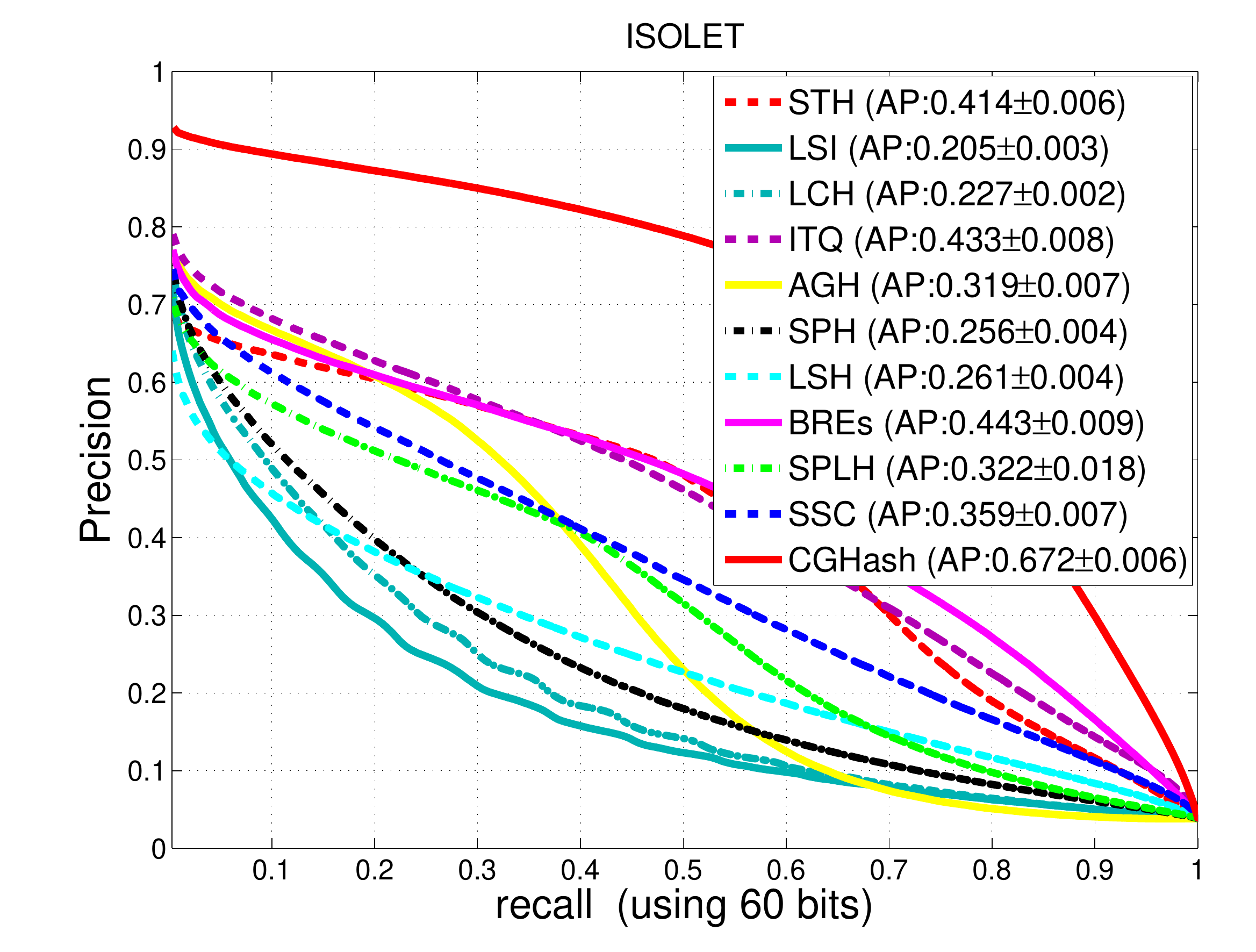}\hspace{-0.03cm}
\includegraphics[scale=0.216]{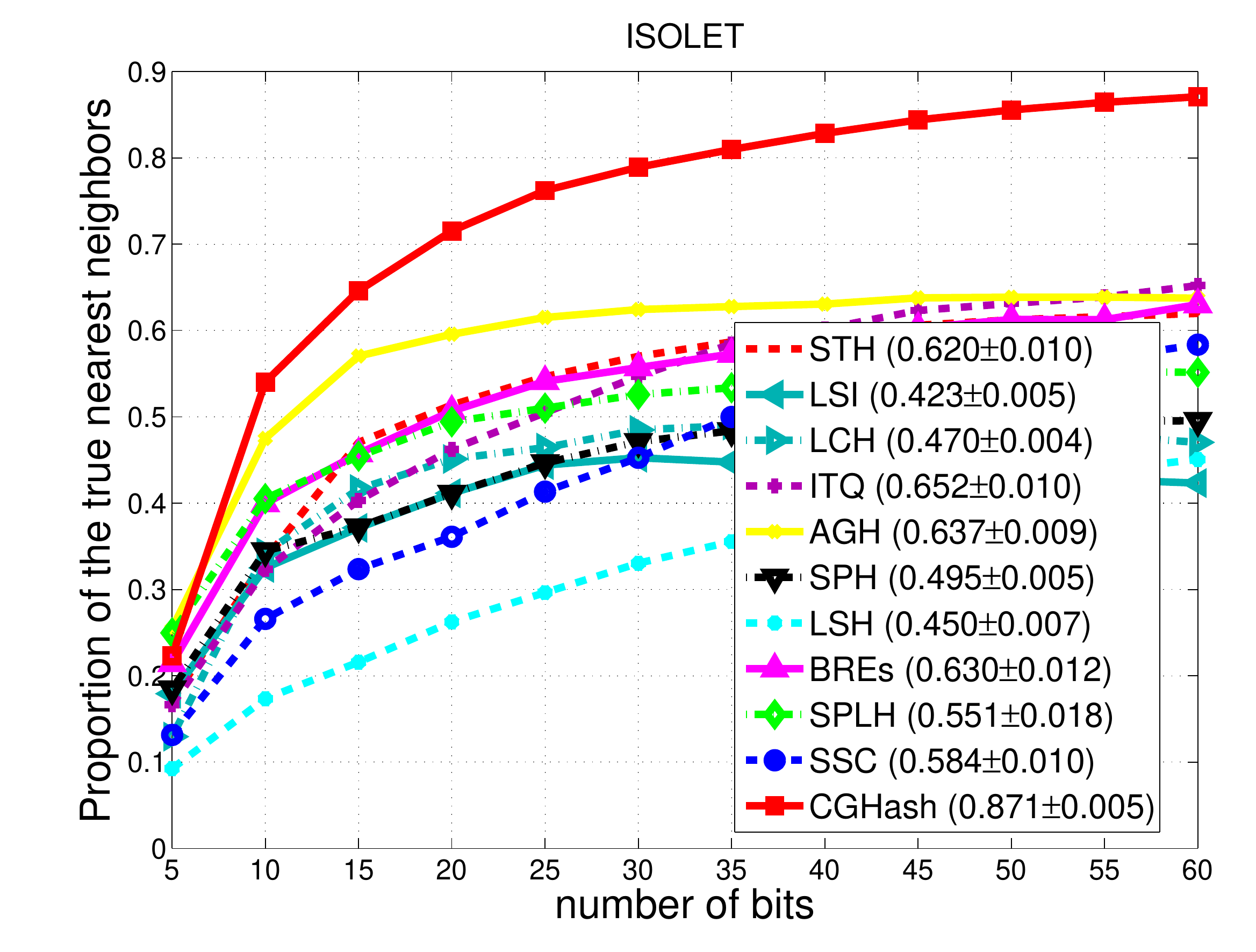} \hspace{-0.08cm}
\includegraphics[scale=0.216]{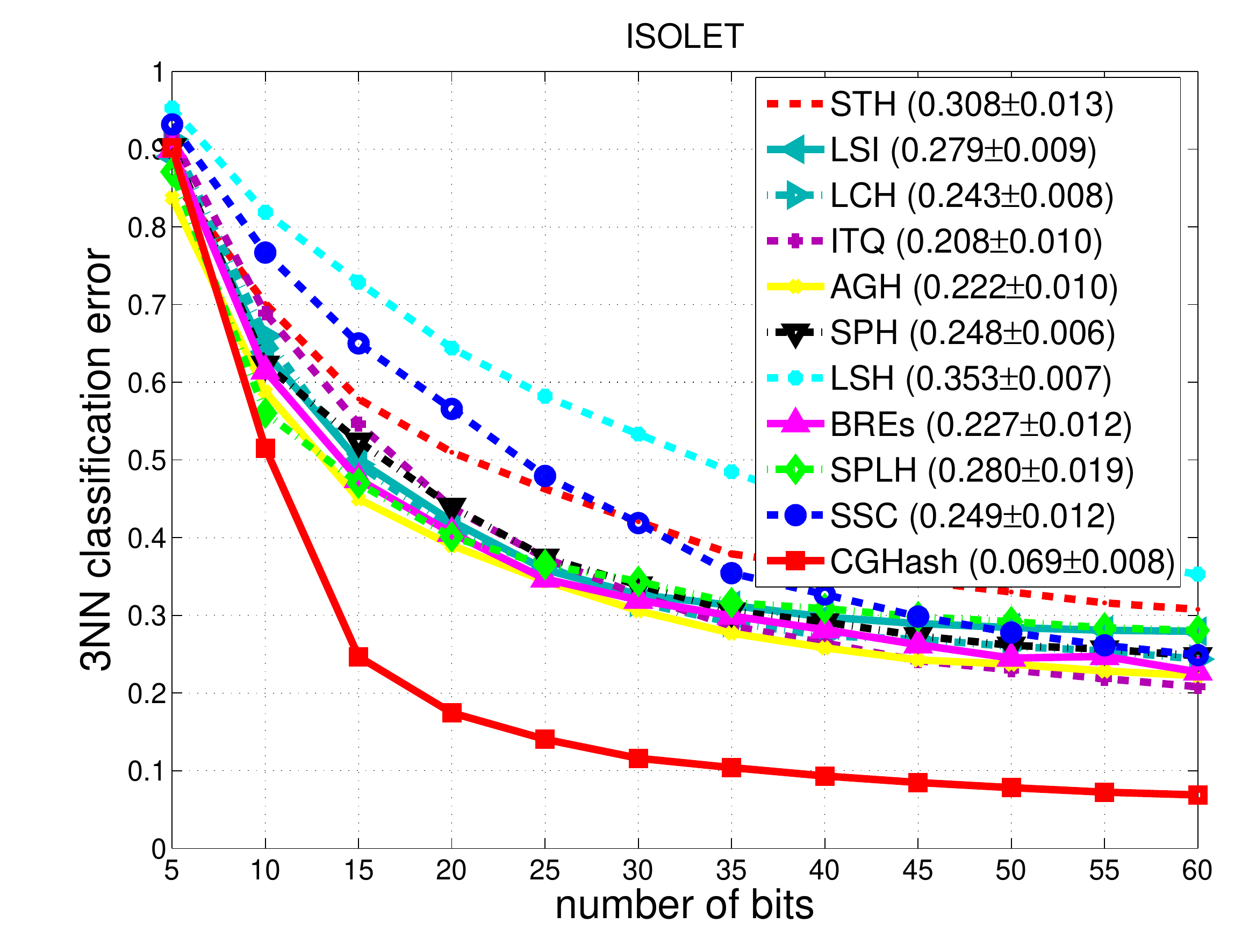}
\caption{The retrieval and classification performances of the proposed
CGHash and 10 other hashing  methods on the ISOLET dataset. The left
plot shows
the average precision-recall performances  using 60 bits.
The middle plot
shows the average performances  using different
code lengths measured as the proportion of the true nearest neighbors with top-50 retrieval.
The right plot shows the average 3-nearest-neighbor classification performances
 using different
code lengths.}
\label{fig:isolet}
\end{figure*}

\begin{figure*}[t]
\centering
\includegraphics[scale=0.216]{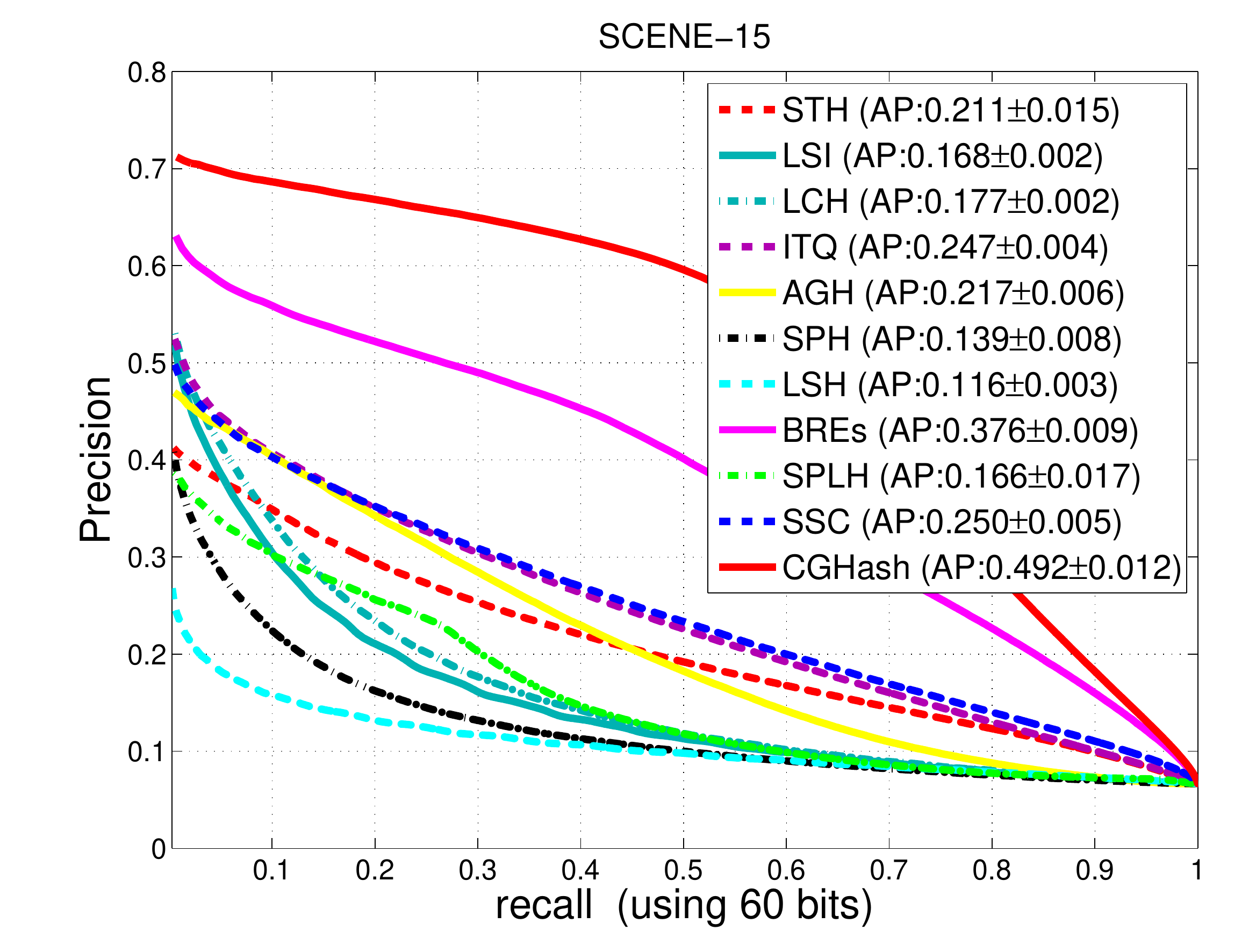}\hspace{-0.03cm}
\includegraphics[scale=0.216]{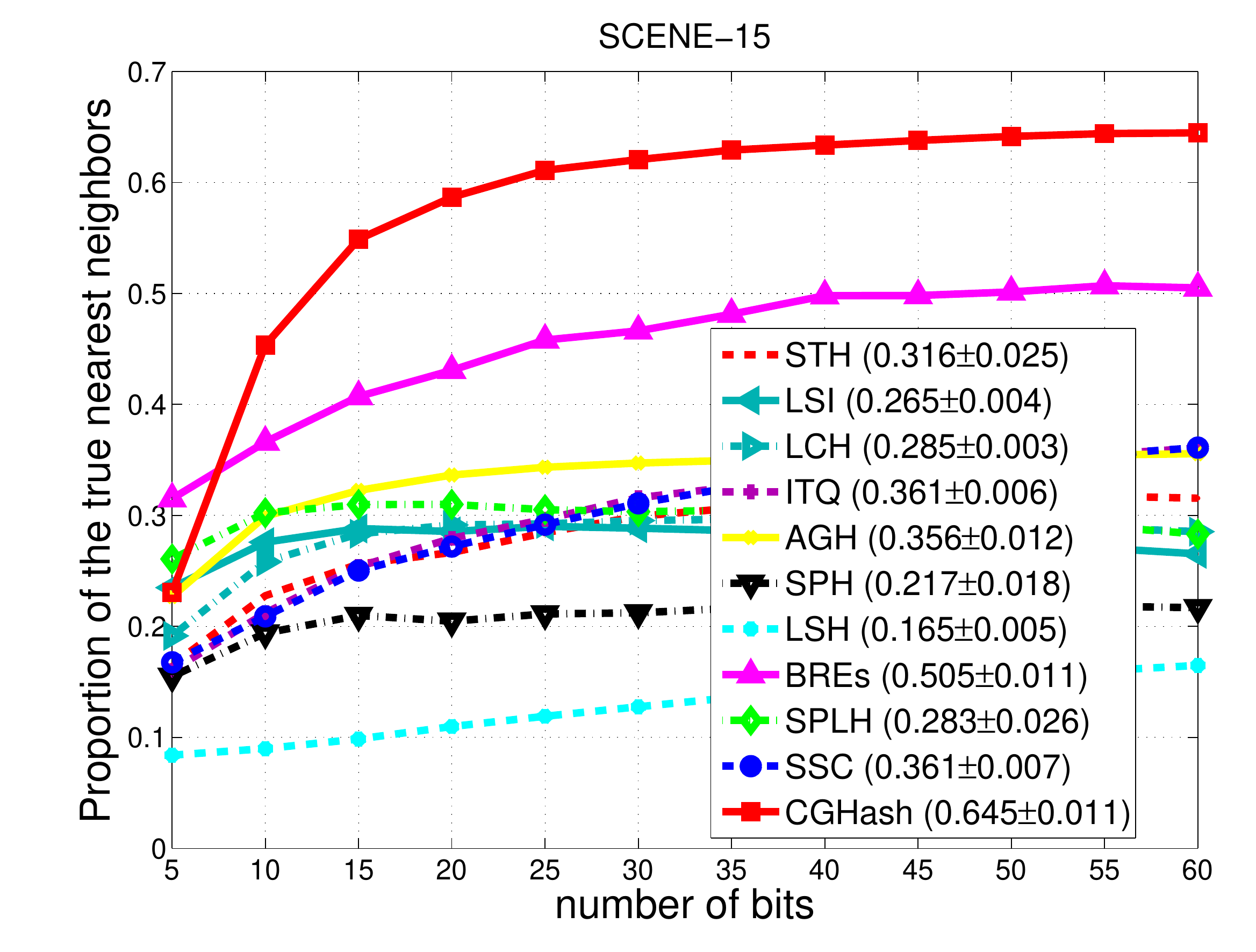} \hspace{-0.12cm}
\includegraphics[scale=0.216]{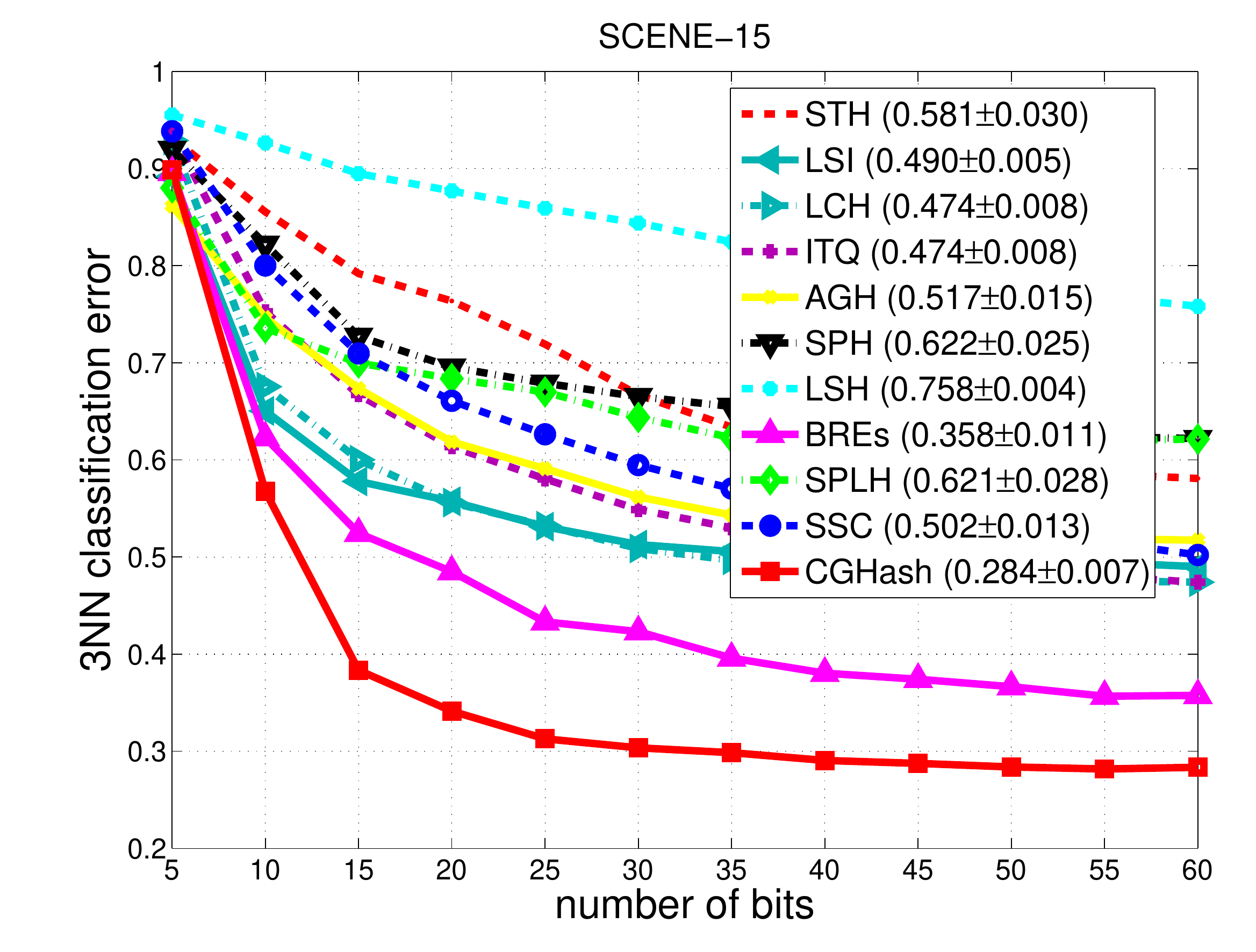}
\vspace{-0.56cm}
\caption{The retrieval and classification performances of the proposed
CGHash and 10 other hashing  methods on the SCENE-15 dataset. 
The description of each plot is the same as in Fig.\ \ref{fig:isolet}.  
}
\vspace{-0.35cm}
\label{fig:scene15}
\end{figure*}

\begin{figure*}[t]
\centering
\includegraphics[scale=0.216]{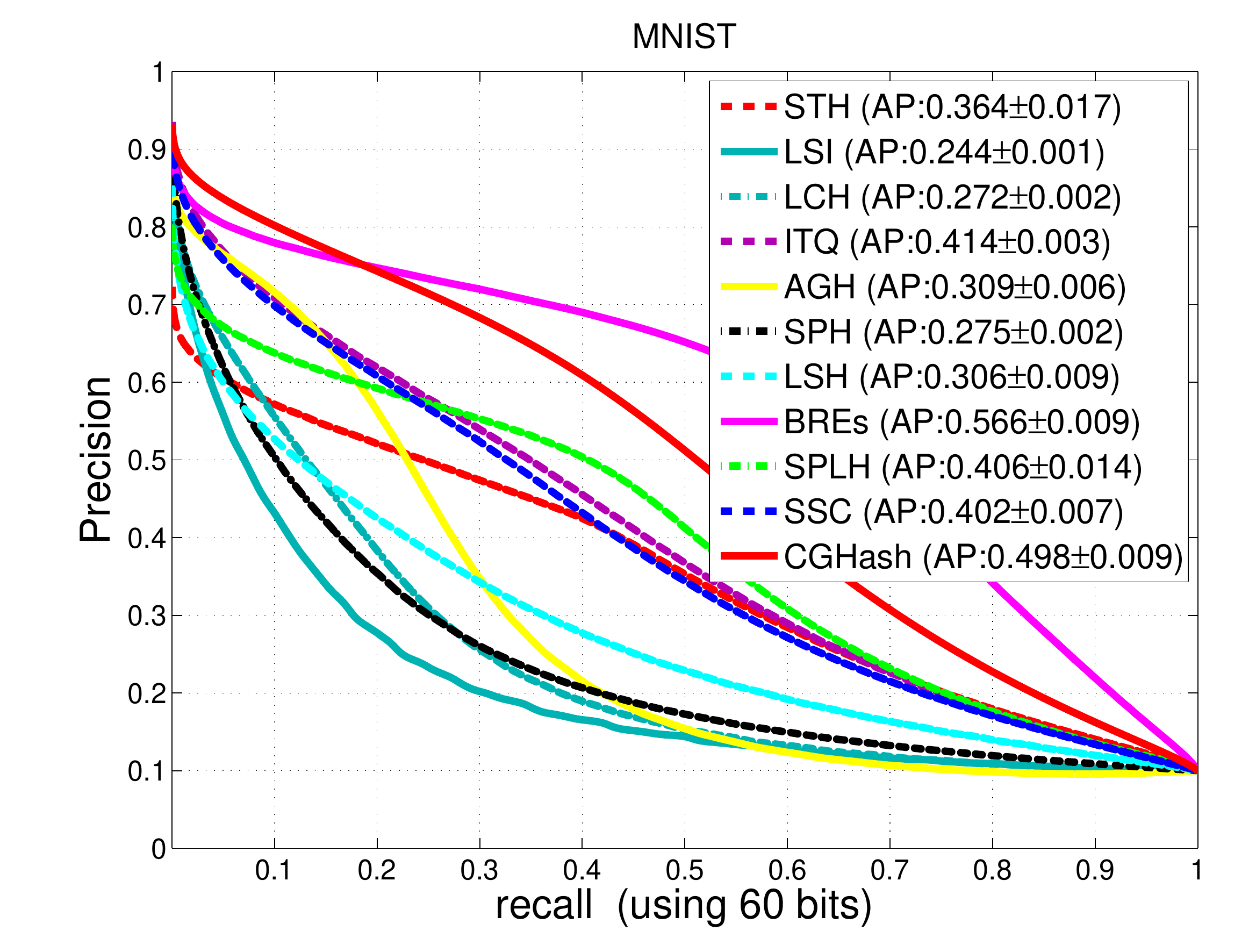} \hspace{-0.08cm}
\includegraphics[scale=0.216]{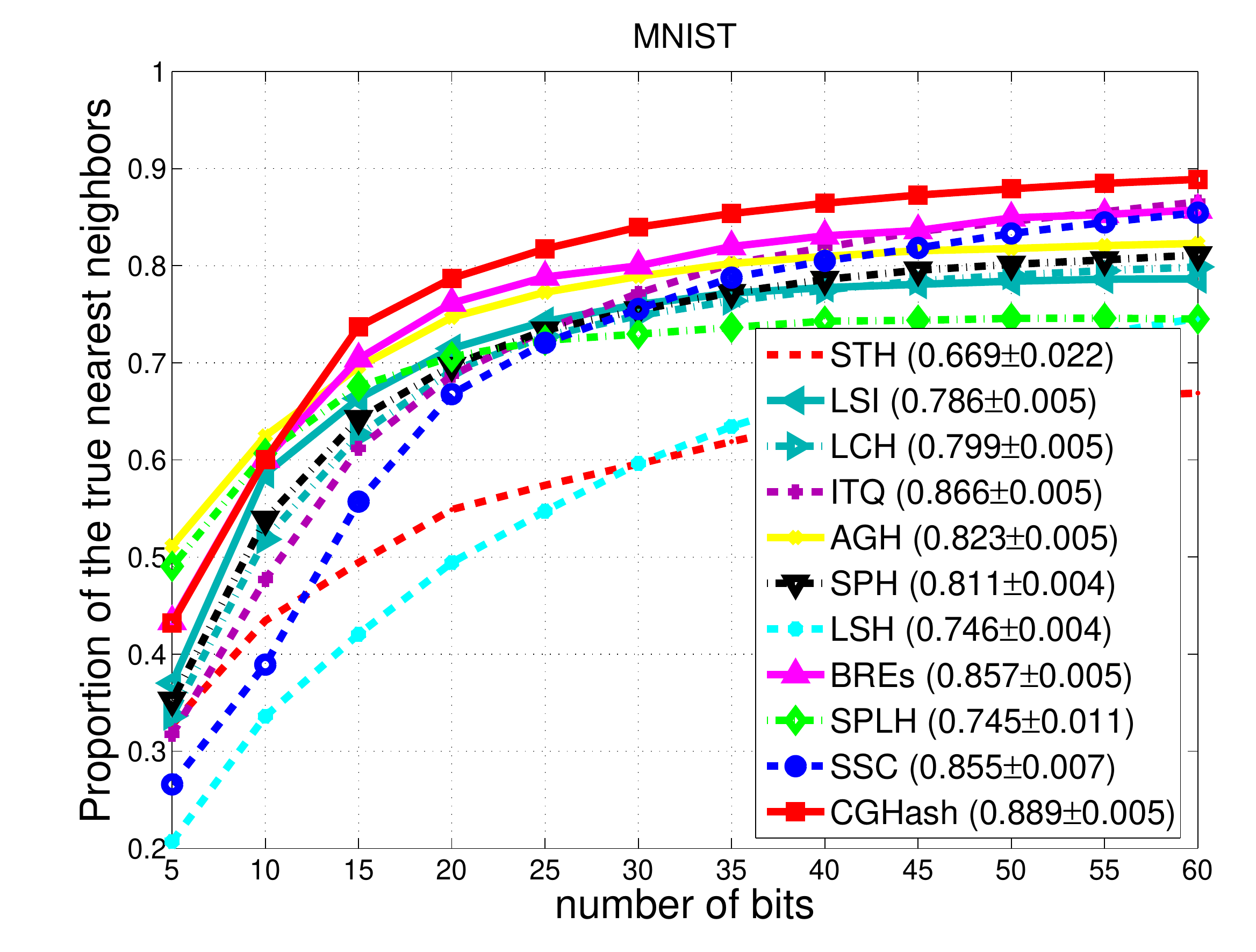} \hspace{-0.08cm}
\includegraphics[scale=0.216]{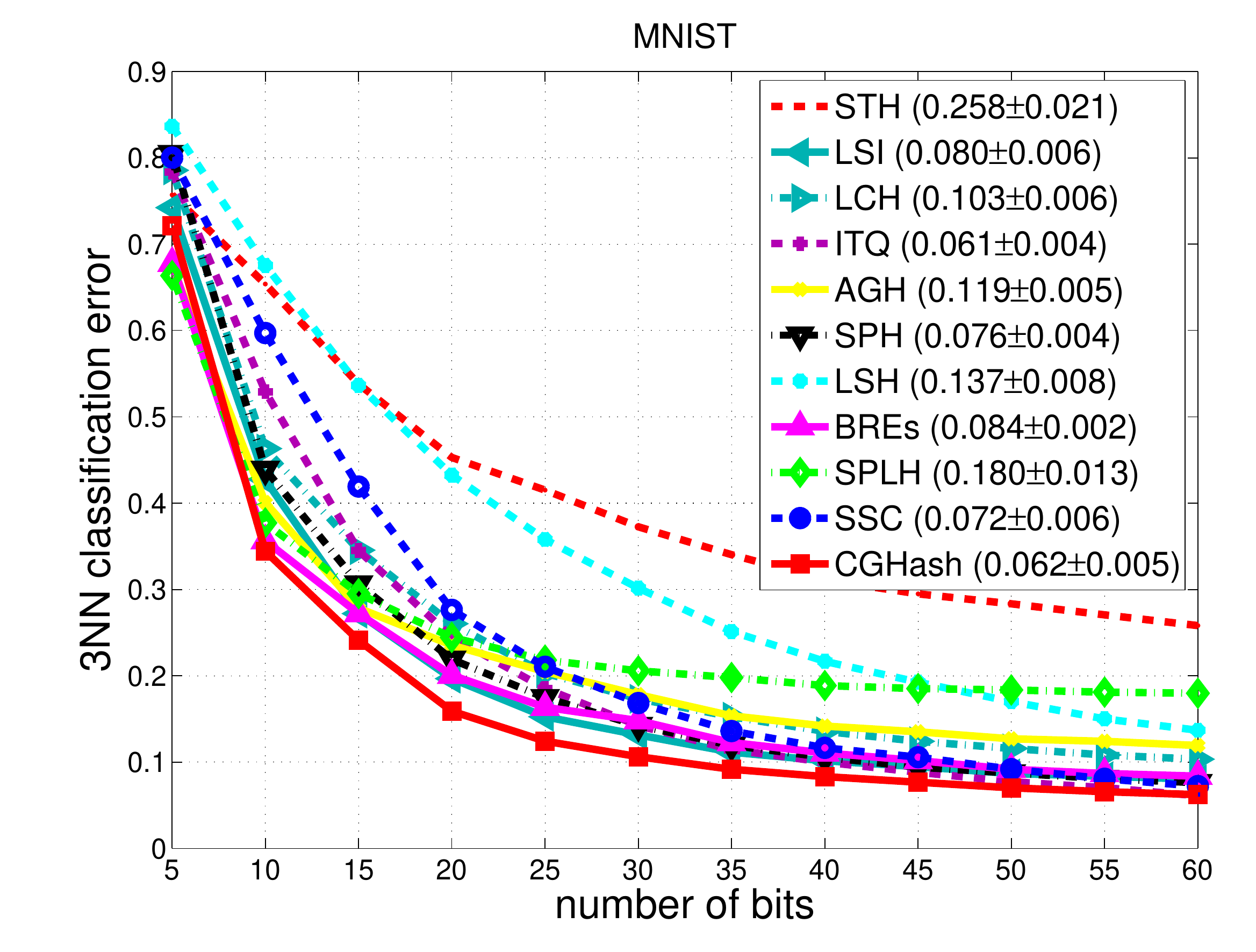}
\vspace{-0.5cm}
\caption{The retrieval and classification performances of the proposed
CGHash and 10 other hashing  methods on the MNIST dataset. 
The description of each plot is the same as in Fig.\ \ref{fig:isolet}. 
}
\vspace{-0.16cm}
\label{fig:mnist}
\end{figure*}

\begin{figure*}[t]
\centering
\includegraphics[scale=0.216]{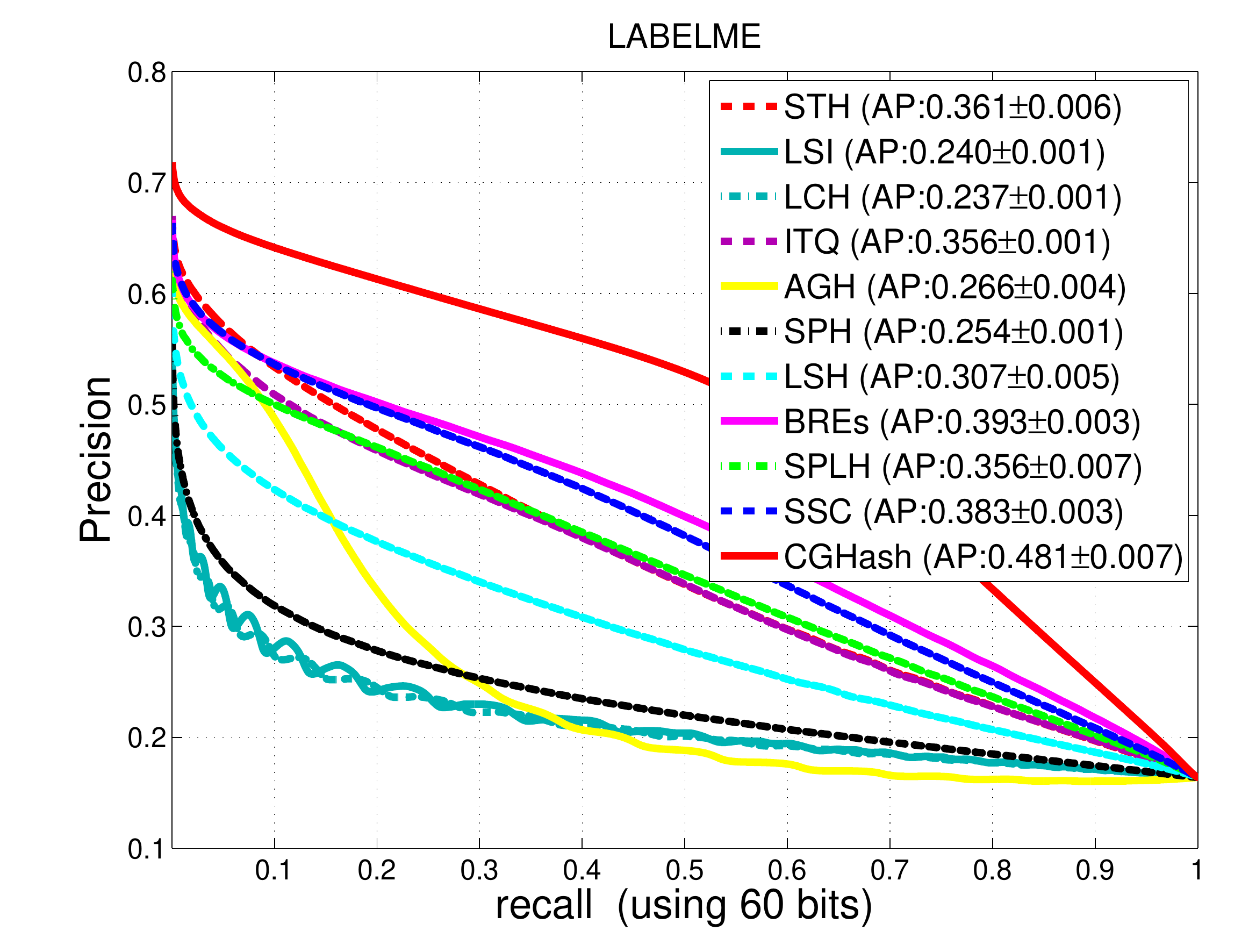} \hspace{-0.08cm}
\includegraphics[scale=0.216]{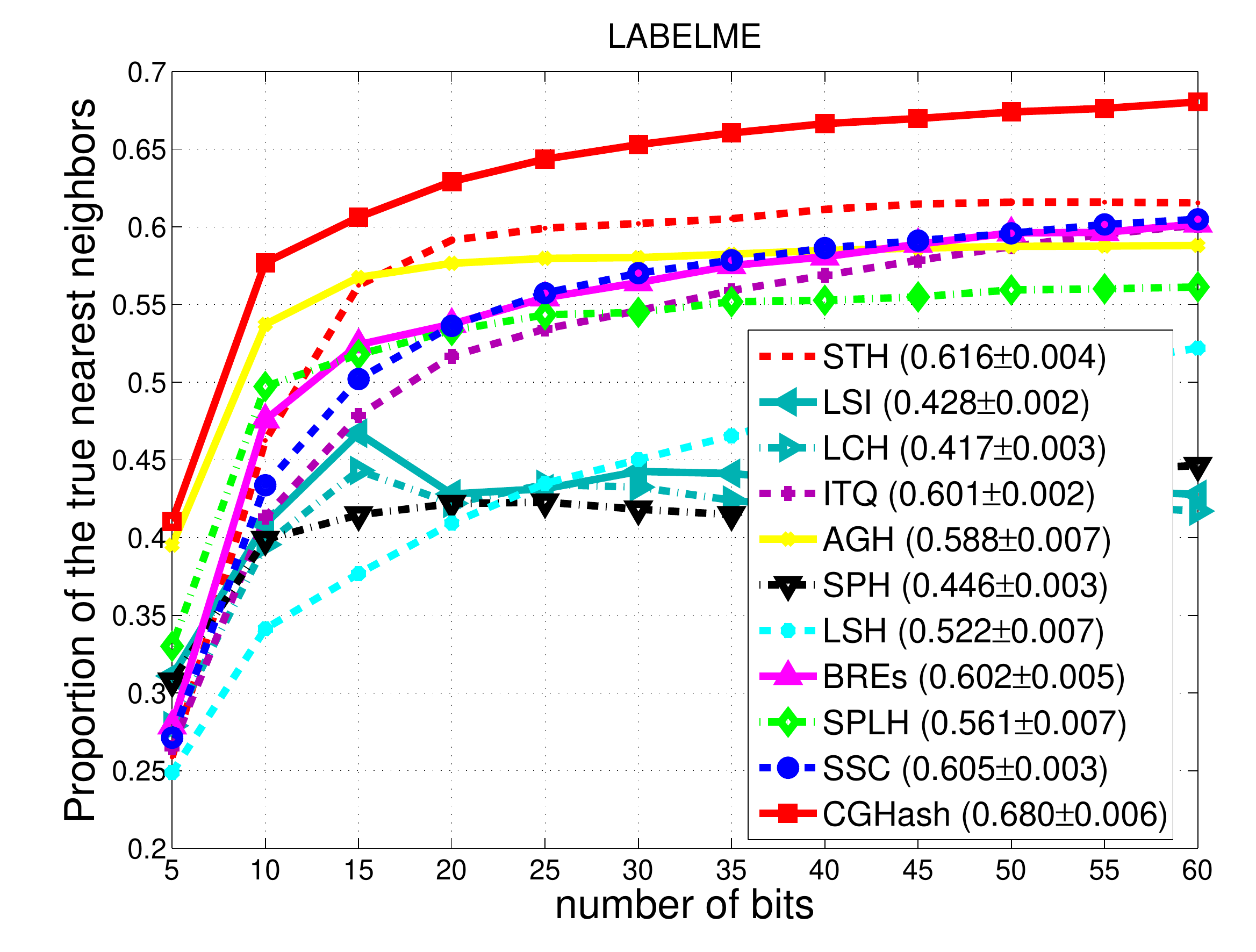} \hspace{-0.08cm}
\includegraphics[scale=0.216]{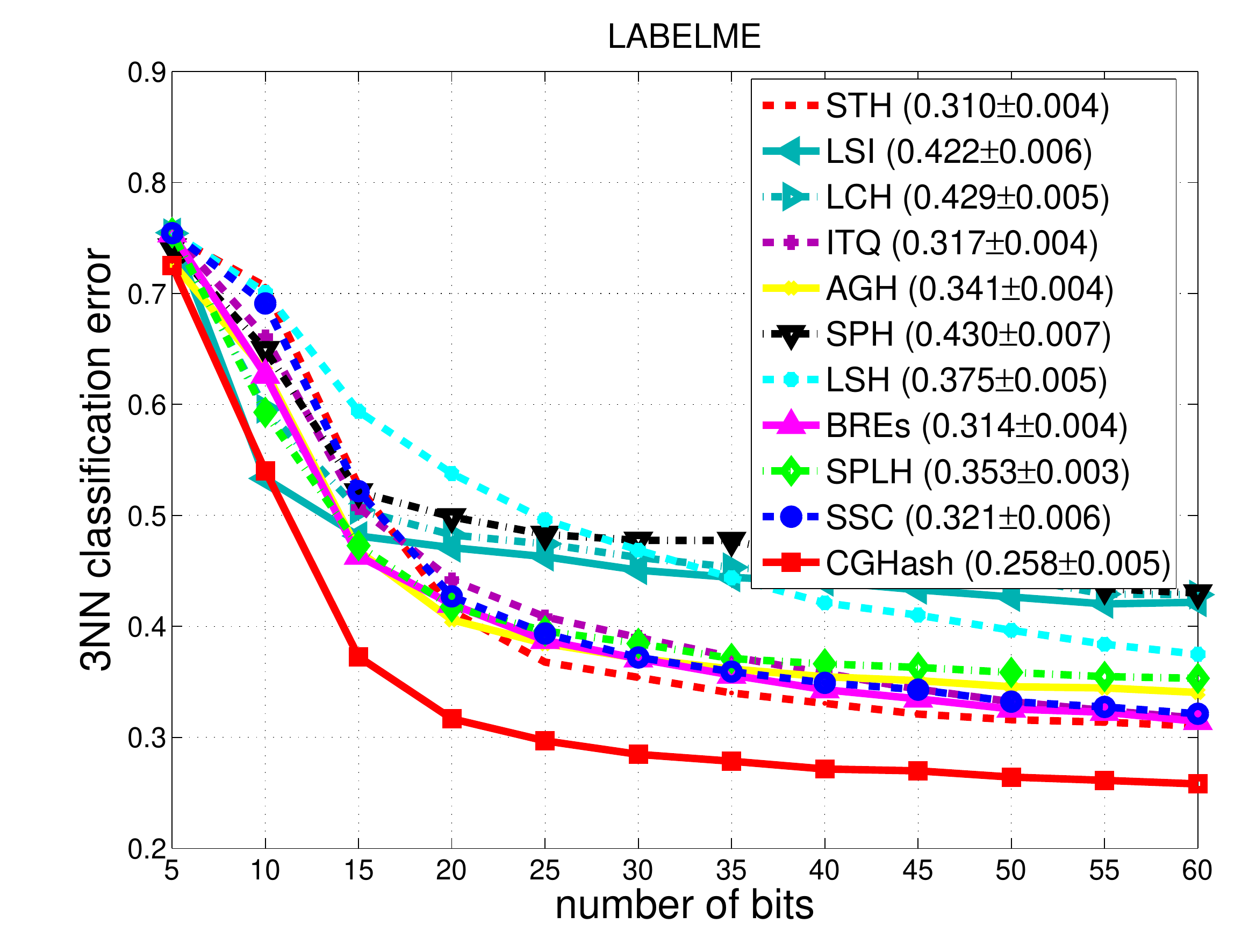}
\vspace{-0.5cm}
\caption{The retrieval and classification performances of the proposed
CGHash and 10 other hashing methods on a subset of the LABELME
dataset.  The description of each plot is the same as in the previous figures.
}
\vspace{-0.26cm}
\label{fig:LABELME}
\end{figure*}

\begin{figure*}[t]
\centering
\includegraphics[scale=0.216]{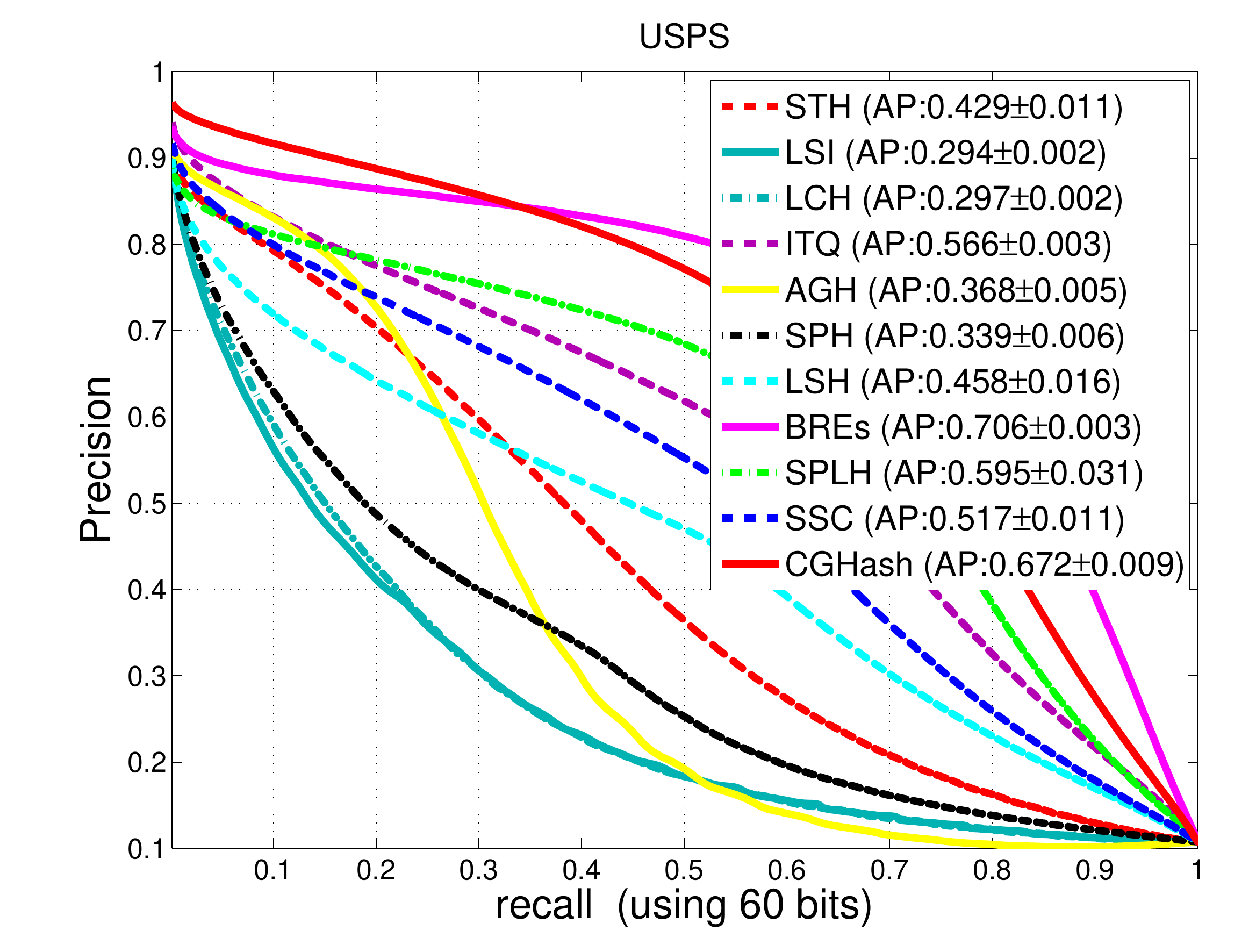} \hspace{-0.08cm}
\includegraphics[scale=0.216]{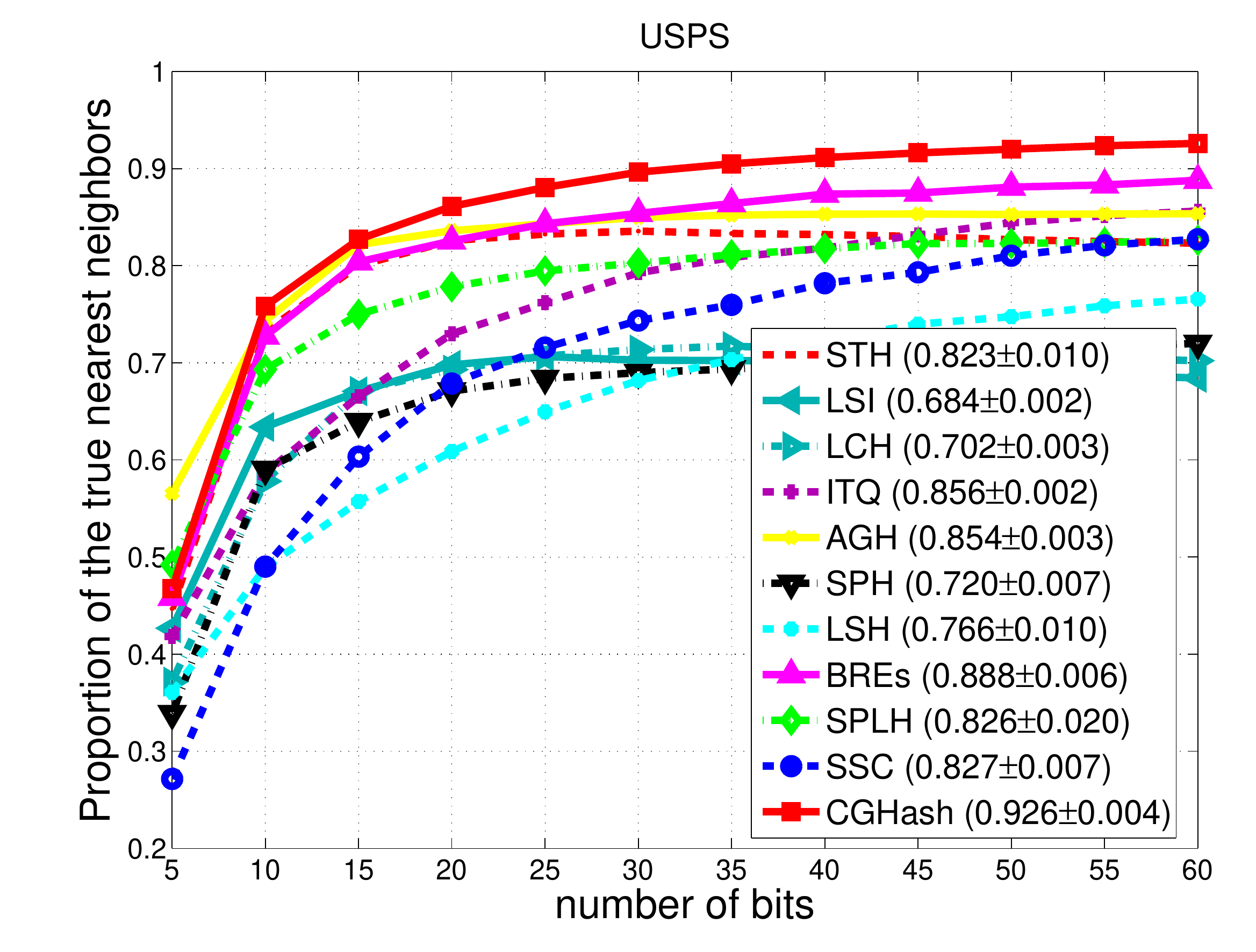} \hspace{-0.08cm}
\includegraphics[scale=0.216]{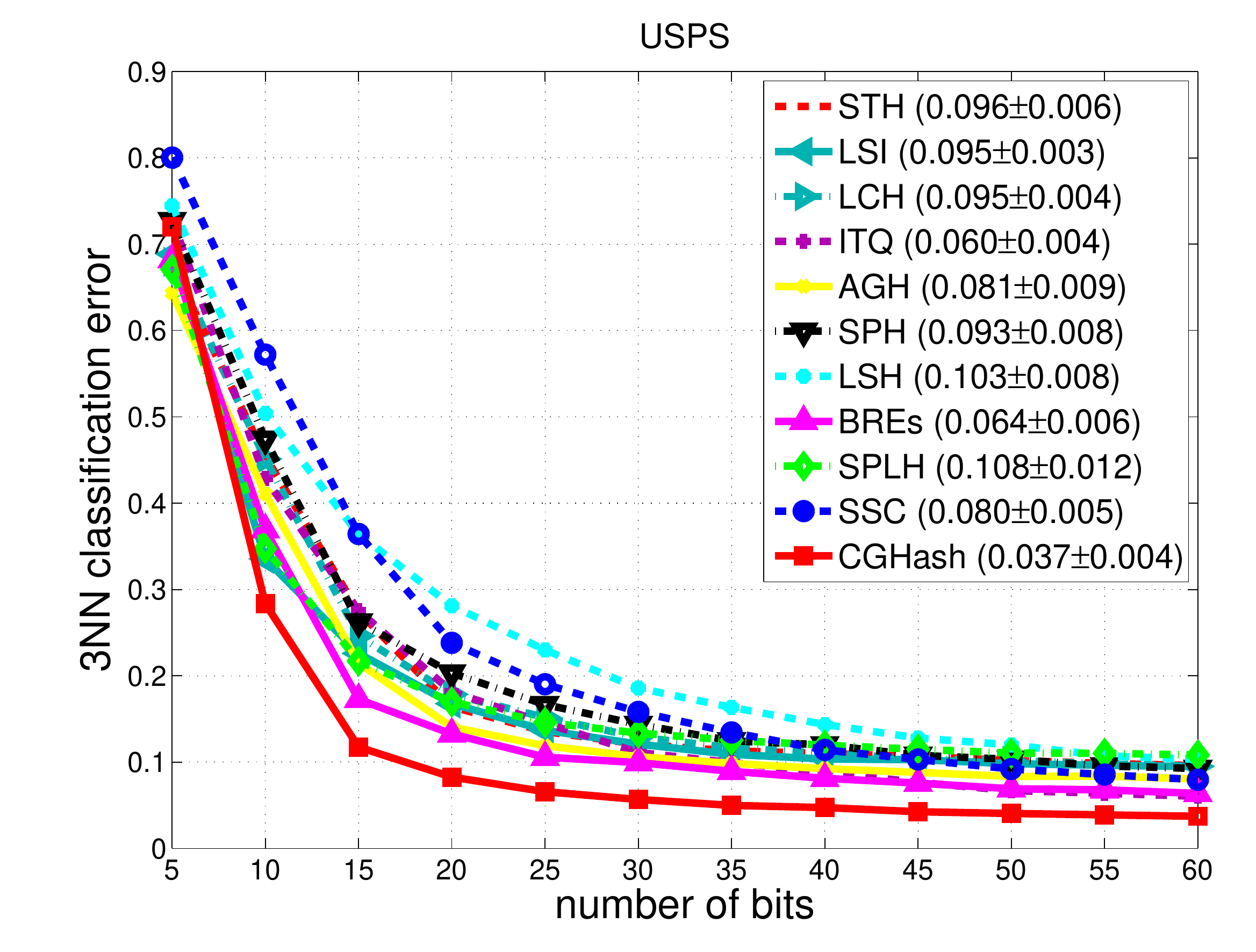}
\vspace{-0.5cm}
\caption{The retrieval and classification performances of the proposed
CGHash and 10 other hashing methods on the USPS dataset. 
The description of each plot is the same as in the previous figures.
}
\vspace{-0.16cm}
\label{fig:USPS}
\end{figure*}

\begin{figure*}[t]
\centering
\includegraphics[scale=0.216]{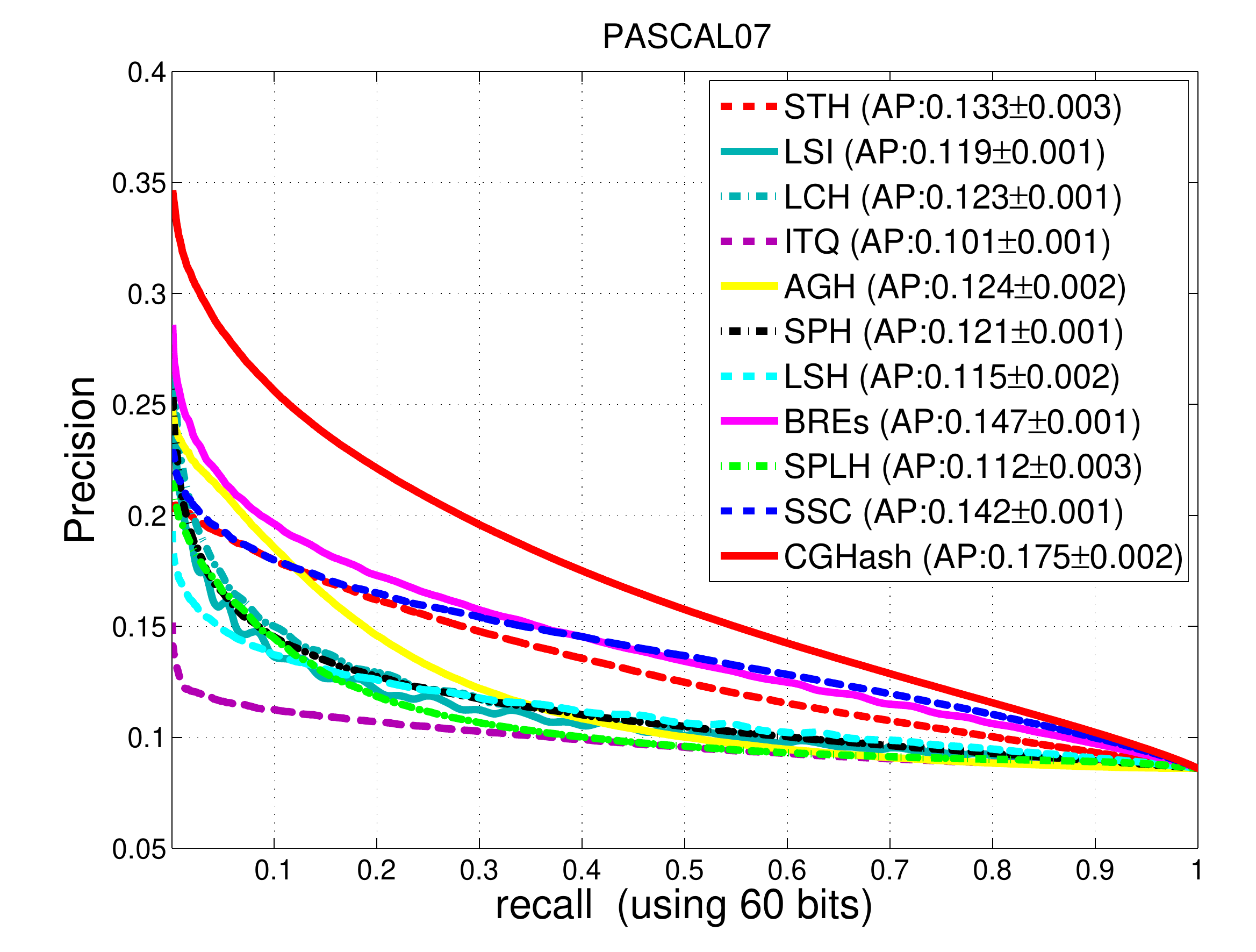}\hspace{-0.03cm}
\includegraphics[scale=0.216]{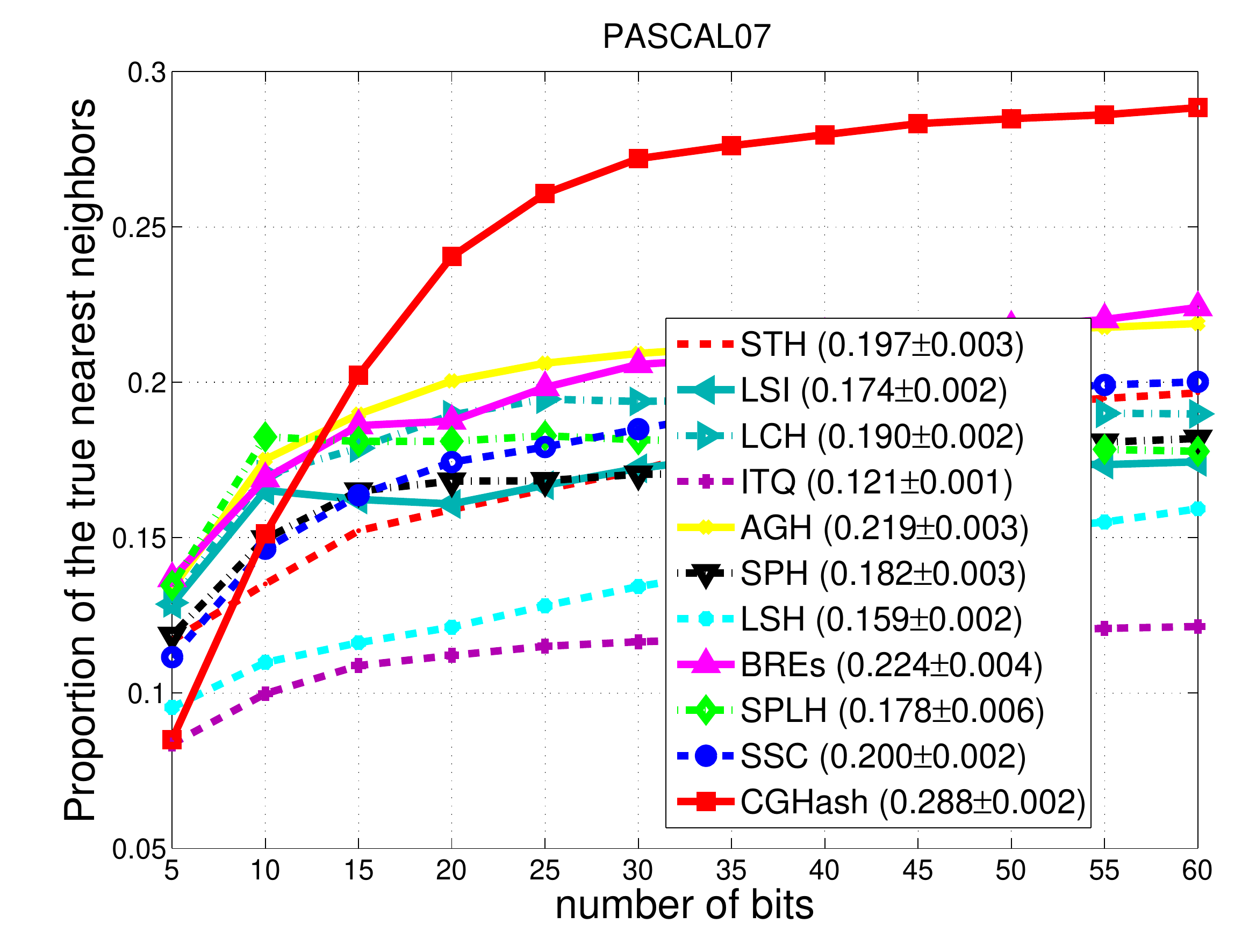}\hspace{-0.08cm}
\includegraphics[scale=0.216]{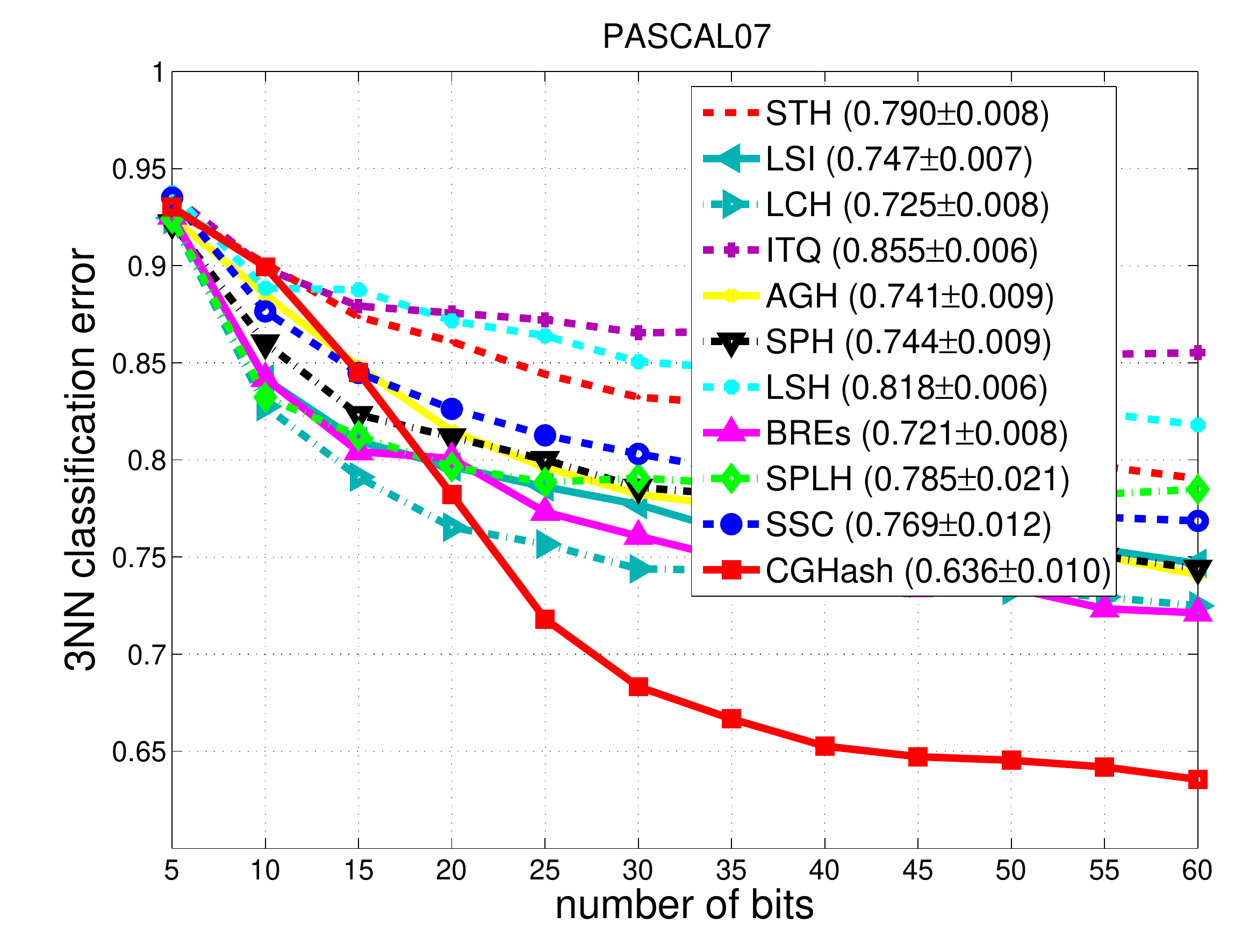}
\vspace{-0.5cm}
\caption{The retrieval and classification performances of the proposed
CGHash and 10 other hashing  methods on the PASCAL07 dataset. 
The description of each plot is the same as in the previous figures.
}
\vspace{-0.26cm}
\label{fig:pascal}
\end{figure*}

\begin{figure*}[t!]
\centering
\includegraphics[scale=0.285]{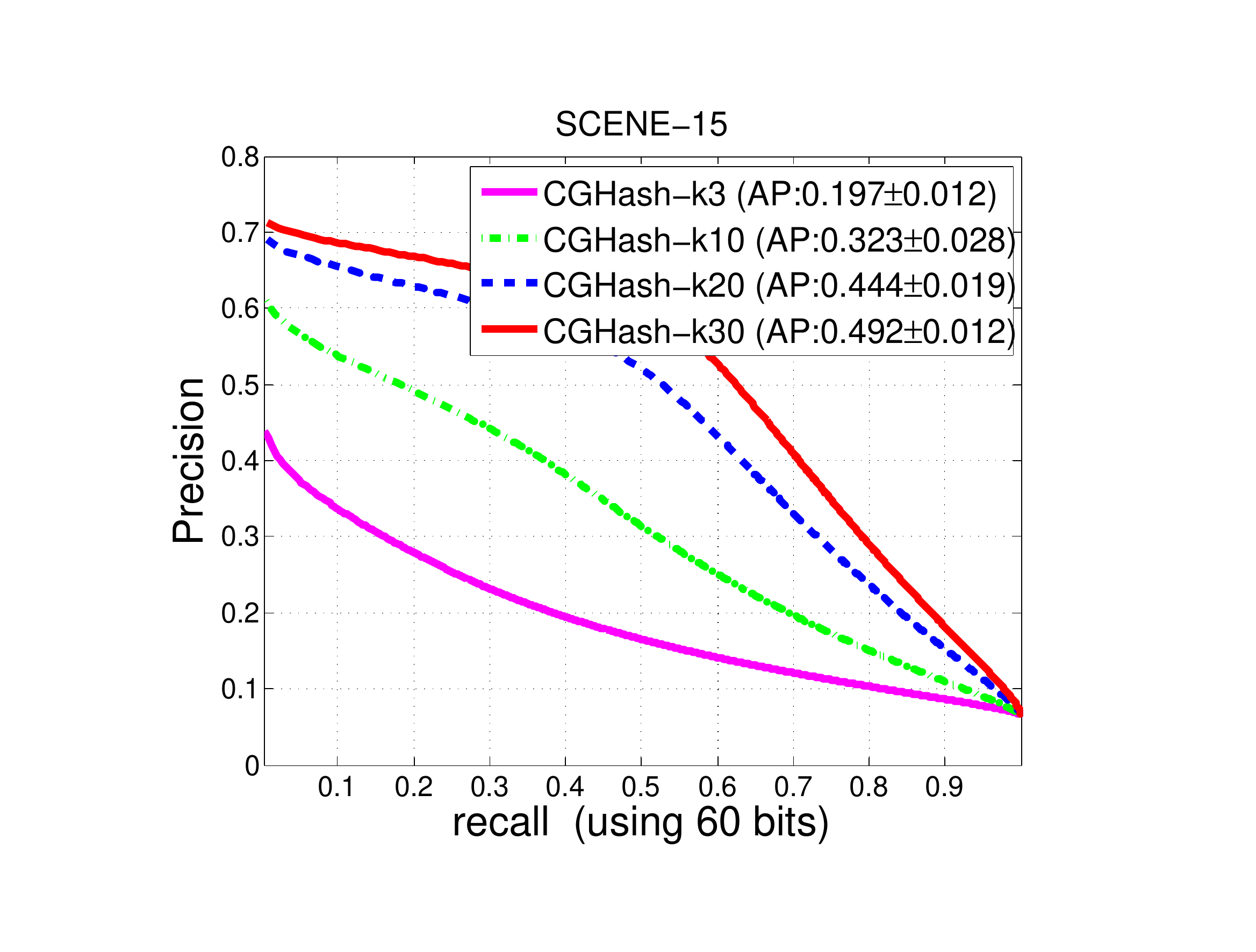} \hspace{0.18cm}
\includegraphics[scale=0.26]{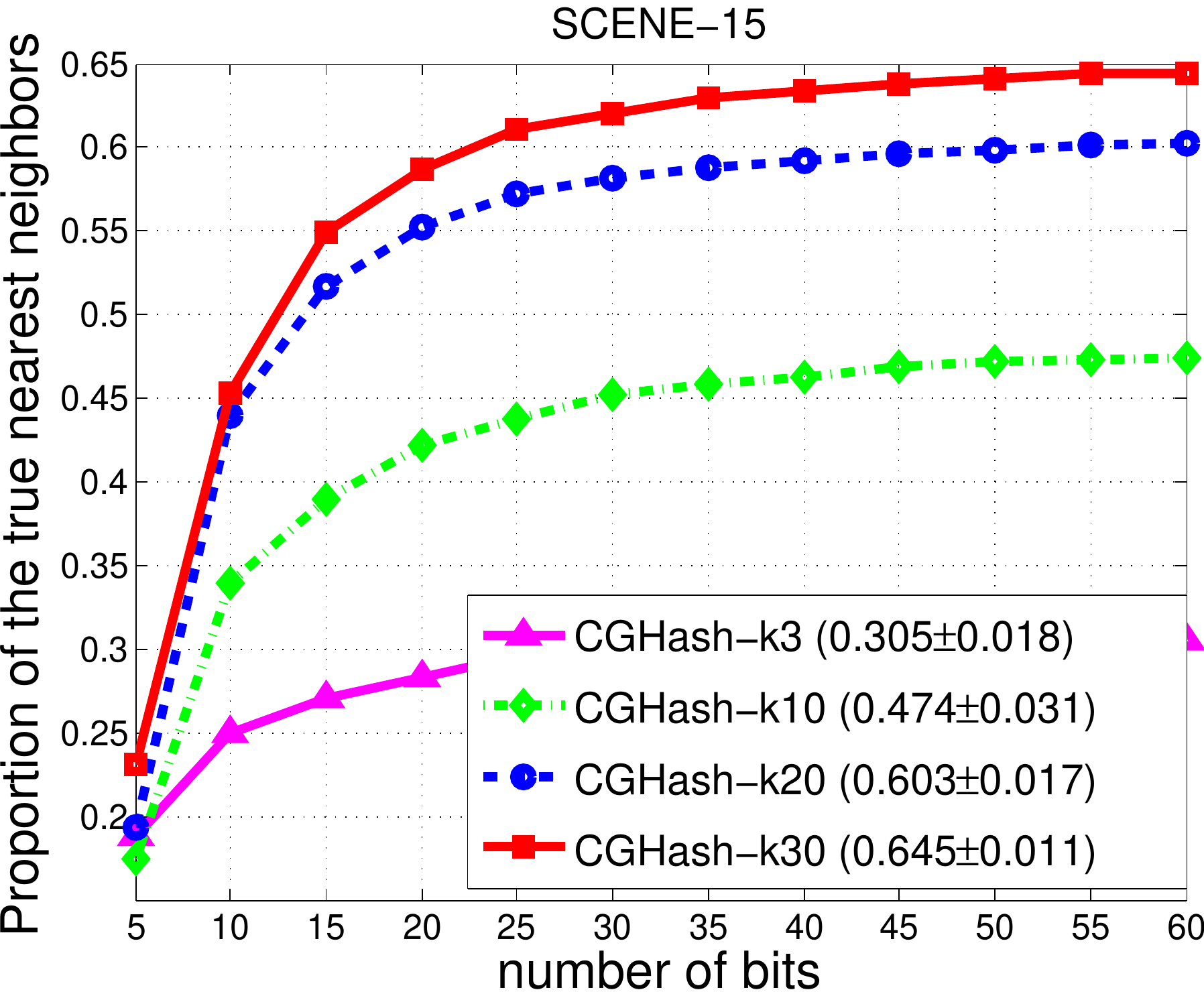} \hspace{0.18cm}
\includegraphics[scale=0.285]{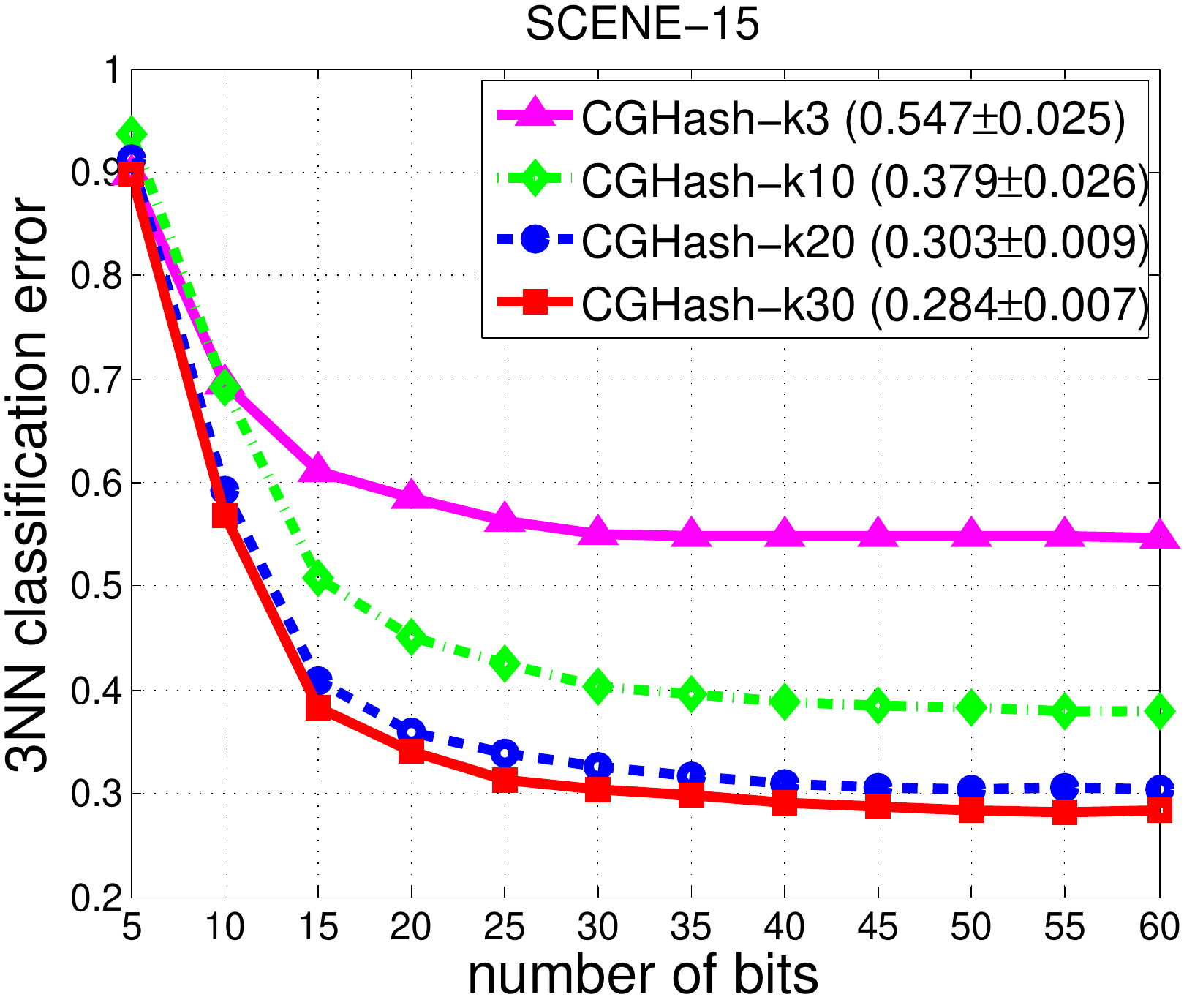}
\vspace{-0.32cm}
\caption{The retrieval and classification performances of the proposed
CGHash using different values of $K$ ($K \in\{3, 10, 20, 30\}$) on the
SCENE-15 dataset. The left plot shows
the average precision-recall performances  using 60 bits.
The middle plot
displays the average performances  using different
code lengths measured as the proportion of the true nearest neighbors with top-50 retrieval.
The right plot shows the average 3-nearest-neighbor classification performances
 using different
code lengths.}
\vspace{-0.16cm}
\label{fig:scene15_comp}
\end{figure*}

    The core idea of column generation is to generate a small subset
    of variables, each of which is sequentially found by selecting the
    most violated dual constraints in the dual optimization
    problem~\eqref{eq:General_Loss_Optimization_dual1B}.  This process
    is equivalent to inserting several primal variables into the
    primal optimization
    problem~\eqref{eq:General_Loss_Optimization_original}.  Here, the
    subproblem for generating the most violated dual constraint (i.e.,
    to find the best hash function) can be defined as:
\begin{align}
&h^{\star}(\cdot)  =  \underset{h(\cdot)\in \mathcal{H}}{\arg\max}
    { \sum}_{i=1}^{|\mathcal{I}|} u_i a^{[i]}
    \notag
    \\
 &   =
    \underset{h(\cdot)\in \mathcal{H}}{\arg\max}
    { \sum}_{i=1}^{|\mathcal{I}|} u_i
    (|h(\bx_{i})-h(\bx^-_{i})|-|h(\bx_{i})-h(\bx^+_{i})|).
    \label{eq:column_generation} \vspace{-0.36cm}
\end{align}
In order to obtain a smoothly differentiable objective function,
we reformulate \eqref{eq:column_generation}
into  the following equivalent form:
\vspace{-0.6cm}
\begin{align}
\notag \\
&
    \underset{h(\cdot)\in \mathcal{H}} {\rm argmax}
    \sum_{i=1}^{|\mathcal{I}|} u_i
    [(h(\bx_{i})-h(\bx^-_{i}))^2-(h(\bx_{i})-h(\bx^+_{i}))^2].
    \label{eq:column_generation3}
\end{align}
The equivalence between \eqref{eq:column_generation} and
\eqref{eq:column_generation3} can be trivially established. 

To globally solve the optimization problem~\eqref{eq:column_generation3} is in
general difficult.
In the case of decision stumps as hash
functions, we can usually exhaustively enumerate all the
possibilities and find the globally best one.
In the case of linear perception as hash functions,
$h(\bx)$ takes the form of
$\mbox{sgn}(\bv^{\T}\bx + \bB)$
where $\mbox{sgn}(\cdot)$ is the sign function.
As a result, the binary hash codes are easily computed by
$(1+h(\bx))/2$.
In practice, we relax $h(\bx)=\mbox{sgn}(\bv^{\T}\bx + \bB)$
to $h(\bx)=\tanh(\bv^{\T}\bx + \bB)$ with
$\tanh(\cdot)$ being the hyperbolic tangent function.
For notional simplicity,
let $\tau_{i+}$ and $\tau_{i-}$ denote
$\tanh(\bv^{\T}\bx_{i} + \bB) - \tanh(\bv^{\T}\bx_{i}^{+} + \bB)$
and $\tanh(\bv^{\T}\bx_{i} + \bB) - \tanh(\bv^{\T}\bx_{i}^{-} + \bB)$, respectively.
Then
we have the following optimization problem:
\begin{align}
& h^{\star}(\cdot) =
\notag \\
&
\underset{h(\cdot)\in \mathcal{H}}{\arg\max}
     \sum_{i=1}^{|\mathcal{I}|} u_i
    [(h(\bx_{i})-h(\bx^-_{i}))^2-(h(\bx_{i})-h(\bx^+_{i}))^2]
\notag \\
 &   = \underset{\bv, \bB}{\arg\max}
    \sum_{i=1}^{|\mathcal{I}|} u_i(\tau_{i-}^{2} - \tau_{i+}^{2}).
    \label{eq:column_generation2}
\end{align}
The above optimization problem can be efficiently
solved by using LBFGS \cite{lbfgs} after feature
normalization.
The initialization of LBFGS can be guided by
LSH~\cite{andoni2006near}.
Namely, we first generate a set of
candidate samples such that
$\bv \sim \mathcal{N}(0, 1)$
and $\bB \sim \mathcal{U}(-1, 1)$
with $\mathcal{N}(\cdot)$
and $\mathcal{U}(\cdot)$
respectively being
the normal and uniform distributions.
Then, we use the
best candidate sample as the initialization that
maximizes the objective function~\eqref{eq:column_generation2}.
In our experiments, we have used
linear perception as hash functions.

{\bf Hashing with $l_\infty$ norm regularization}
We show here that we can also use other regularization terms such as
the $l_\infty$ norm.
With the $l_\infty$ norm regularization, the  primal problem
is defined as:
\begin{equation}
\underset{\bomega, \brho}{\min}
{\textstyle \sum}_{i=1}^{|\mathcal{I}|}  \,  f(\rho_{i}) +
C\|\bomega\|_{\infty},
\; \mbox{s.t.}  \;  \bomega \succeq \bzero; \rho_{i} =
\ba_{i}^{\T}\bomega, \thickspace \forall i.
\label{eq:General_Loss_Optimization_original_infinity_norm} %
\end{equation}
This optimization problem is equivalent to:
\begin{equation}
\underset{\bomega, \brho}{\min}  \;
    {\textstyle \sum}_{i=1}^{|\mathcal{I}|}
    f(\rho_{i}),
    \,
\mbox{s.t.} \;  \|\bomega\|_{\infty} \leq  C'; \bomega\succeq \bzero; \thinspace \rho_{i} = \ba_{i}^{\T}\bomega,\forall i,
\label{eq:General_Loss_Optimization_original_infinity_norm_relaxed}
\end{equation}
where $C'$ is a properly selected constant, related to $ C $ in
\eqref{eq:General_Loss_Optimization_original_infinity_norm}.
Due to $\bomega \succeq \bzero$ and $\|\bomega\|_{\infty}\leq C'$, we obtain
$\bzero \preceq \bomega \preceq C'\bone$.
Therefore, the Lagrangian can be written as: %
\begin{equation*}
    {L} =  \sum_{i=1}^{|\mathcal{I}|}f(\rho_{i}) +
\bq^{\T}\bomega - C'\bq^{\T}\bone -\bp^{\T}\bomega +
   \sum_{i=1}^{|\mathcal{I}|}u_{i}(\ba_{i}^{\T}\bomega - \rho_{i}), %
\end{equation*}
where $\bp$, $\bq$, $\bu$ are Lagrange multipliers.
 Similar to the $ \ell_1 $ norm case,  we can easily derive the dual problem  as:
\begin{equation}
    \underset{\bu, \bq}{\min} \; {\textstyle \sum}_{i=1}^{|\mathcal{I}|}f^{\ast}(u_i) +
C' \bone ^\T \bq,
\;
\mbox{s.t.} \; A\bu \succeq -\bq.
\label{eq:general_loss_infinity} \vspace{-0.12cm}
\end{equation}
By reversing the sign of $\bu$, we can reformulate \eqref{eq:general_loss_infinity}
as its equivalent form:
\begin{equation}
\underset{\bu, \bq}{\min}
\; {\textstyle \sum}_{i=1}^{|\mathcal{I}|}f^{\ast}(-u_i) + C' \bone
^\T \bq, \;
\mbox{s.t.} \; A\bu \preceq \bq.
\label{eq:General_Loss_Optimization_dual1B_infinity} \vspace{-0.26cm}
\end{equation}

The KKT condition in this $ l_\infty$ regularized case is the same as
\eqref{EQ:KKT1}.
Also the rule to generate the best hash function (i.e., the most violated
constraint in \eqref{eq:General_Loss_Optimization_dual1B_infinity})
remains the same as in the $ l_1 $ norm case that we have discussed.
Note that both the primal problems
\eqref{eq:General_Loss_Optimization_original} and
\eqref{eq:General_Loss_Optimization_original_infinity_norm_relaxed}
can be efficiently solved using quasi-Newton methods such as L-BFGS-B
\cite{lbfgs} by eliminating the auxiliary variable $ \brho$.

{\bf Extension}
     To demonstrate the flexibility of the proposed framework, we
     show an example that considers an addition pairwise
     information.
    Assume that we have information about a set of duplets
    that they are neighbors to each other or they are from the same
    class.  So the distance between these duplets should be
    minimized. We can easily include such a term in our objective
    function. Formally, let us denote the duplet set as
    $  {\cal D} = \{  (  \bx_k,  \bx_k^+  )  \}  $
    and we want to minimize the divergence
    $  \sum_{k=1}^{ | \cal D|  }
                d_{\cal H} (  \bx_k, \bx^+_k  )
        = \sum_j  \omega_j (
        \sum_{k}   | h_j( \bx_k ) - h_j( \bx_k^+ )|)
        = \sum_j s_j  \omega_j
    $
    with $ s_j = \sum_{k=1}^{ |\cal D| }
    $ $  | h_j( \bx_k ) $ $ - h_j( \bx_k^+ )| $
    being a nonnegative constant given $ h_j( \cdot ) $.
    If we use this term to replace the $ l_1 $ regularization term
    $ \sum_j \omega_j $ in
    the primal \eqref{eq:General_Loss_Optimization_original},
    all of our analysis still holds and   Algorithm
    \ref{alg:boosting_hash} is still applicable with minimal
    modification, because the new term can be simply seen as
    a weighted $ l_1 $ norm.

\section{Experimental results}

{\bf Experimental setup} In order to evaluate the proposed column
generation hashing method (referred to as CGHash), we have conducted a
set of experiments on six benchmark datasets. %
To train data-dependent hash functions, each dataset is randomly split into
a training subset and a testing subset.
This training/testing split is repeated 5 times, and the average
performance over these 5 trials is reported here.

In the experiments, the proposed hashing method
is implemented by using the squared hinge loss function
with the $l_1$ regularization norm (as shown in the supplementary file).
Moreover, the triplets used for learning hash functions
are generated in the same way as~\cite{weinberger2006distance}.
Specifically, given a training sample, we select the
$K$ nearest neighbors from its associated same-label training samples
as relevant samples,
and then choose the
$K$ nearest neighbors from its associated different-label training samples
as irrelevant samples ($K=30$ for the SCENE-15 dataset and $K=10$ for the other datasets).
The trade-off control factor $C$ is
cross-validated. We found
that, in a wide range, the trade-off control factor $C$
does not have a significant impact on the performance.

{\bf Competing methods}
To demonstrate the effectiveness of the proposed hashing method (CGHash), we
compare with some other state-of-the-art hashing methods 
 quantitatively.
For simplicity, they are respectively referred to as
LSH (Locality Sensitive Hashing~\cite{andoni2006near}),
SSC (Supervised Similarity Sensitive Coding~\cite{torralba2008small} as a modified version of \cite{shakhnarovich2003fast}),
LSI (Latent Semantic Indexing~\cite{deerwester1990indexing}),
LCH (Laplacian Co-Hashing~\cite{zhang2010laplacian}),
SPH (Spectral Hashing~\cite{weiss2008spectral}), STH (Self-Taught hashing~\cite{zhang2010self}),
AGH (Anchor Graph Hashing~\cite{liu2011hashingGraphs}),
BREs (Supervised Binary Reconstructive Embedding~\cite{kulis2009learning}),
SPLH (Semi-Supervised Learning Hashing~\cite{wang2010semi}),
and ITQ (Iterative Quantization~\cite{gong2012iterative}).
Making a comparison with the above competing methods
can verify the effect of learning hashing functions and
show the performance differences in the context of hashing methods.

{\bf Evaluation criteria} For a quantitative performance comparison, we introduce the following three evaluation criteria:
i) precision-recall curve;
ii) proportion of true neighbors in top-$k$ retrieval; and iii) $K$-nearest-neighbor classification.
In the experiments, the aforementioned retrieval performance scores are
averaged over all test queries in the dataset.
For i), the precision-recall curve
is computed as follows: $\mbox{precision} = \frac{\# \mbox{retrieved  relevant sampels}}{\# \mbox{all retrieved samples}}$
and $\mbox{recall} = \frac{\# \mbox{retrieved  relevant sampels}}{\# \mbox{all relevant samples}}$.
For ii), the proportion of true neighbors in top-$k$ retrieval is
calculated as: $\frac{\mbox{\#retrieved true neighbors}}{k}$.
For iii), each test sample is classified by a majority
voting in $K$-nearest-neighbor classification.

{\bf Quantitative comparison results}
Figs.~\ref{fig:isolet}--\ref{fig:pascal} show the retrieval and classification performances
of all the hashing methods using different code lengths on the six datasets.
In each of these figures, we report quantitative comparison
results of all the hashing methods in the following three
aspects:
1) the average precision-recall performances using the maximum code length,
and the average precisions together with standard deviations (as shown in the legend of each figure);
2) the average performances using different
code lengths in the proportion of the true nearest neighbors with top-50 retrieval,
and the average proportion results together with their standard deviations
in the case of the maximum code length (as shown in the legend of each figure);
and 3) the average $K$-nearest-neighbor classification performances
using different
code lengths, and the average classification results together with their standard deviations
in the case of the maximum code length (as shown in the legend of each figure).

From Figs.~\ref{fig:isolet}--\ref{fig:pascal}, we clearly see that the
proposed CGHash obtains the larger areas under the
precision-recall curves than the competing hashing methods.
In addition,
we observe that CGHash achieves the higher
proportions of the true nearest neighbors with top-50 retrieval at most times.
Moreover,
it is seen that CGHash
has lower classification errors than the competing methods
in most cases.

Fig.~\ref{fig:scene15_comp} shows
the retrieval and classification performances
of the proposed CGHash using different values of
$K$ on the SCENE-15 dataset.
It is seen from Fig.~\ref{fig:scene15_comp}
that in general the performance  is improved
as $K$ increases.

\begin{figure}[t]
\centering
\includegraphics[scale=0.16]{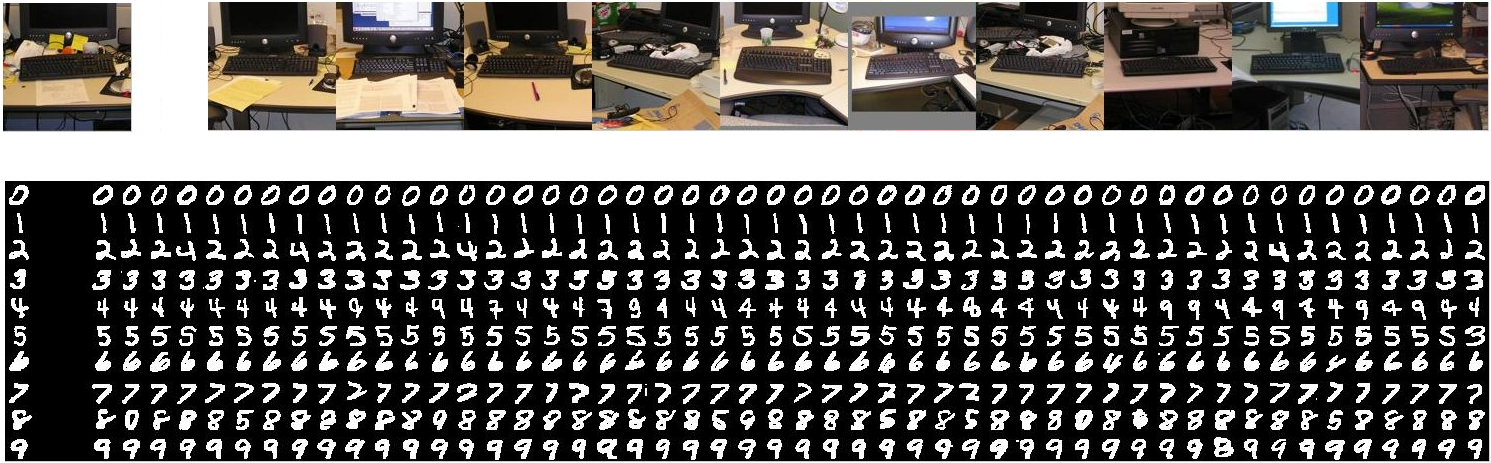}
\vspace{-0.6cm}
\caption{Two retrieval examples for CGHash on the LABELME and MNIST datasets.
The left part shows query samples while the right part displays the
first a few nearest neighbors obtained using CGHash.}
\vspace{-0.36cm}
\label{fig:retrieval_example}
\end{figure}

Besides, Fig.~\ref{fig:retrieval_example}
shows two
retrieval examples on the
MNIST and LABELME datasets.
From Fig.~\ref{fig:retrieval_example},
we observe that CGHash
obtains the visually accurate nearest-neighbor-search
results.

\textbf{Conclusion}
    We have proposed a novel hashing  method that is implemented using
    column generation-based convex optimization. By taking into
    account a set of constraints on the triplet-based relative
    ranking, the proposed hashing  method is capable of learning
    compact hash codes.
    Such a set of constraints are incorporated into the large-margin
    learning framework.
    Hash functions are then learned iteratively using column
    generation.
    Experimental results on several datasets have shown that the
    proposed hashing  method achieves improved performance compared
    with state-of-the-art hashing methods in nearest-neighbor
    classification, precision-recall, and proportion of true nearest
    neighbors retrieved.

    {\it
    This work is in part supported by 
    ARC grants LP120200485
    and  FT120100969.
    }

\bibliographystyle{icml2013}

\end{document}